\documentclass{article} 
\usepackage{style/custom_iclr2021_base}
\usepackage{style/neurips_2020}
\usepackage{times}
\usepackage{booktabs}
\usepackage{caption}
\usepackage{natbib}
\usepackage{amsmath,amsfonts,bm}

\def\eqref#1{equation~\ref{#1}}

\def\1{\bm{1}}

\DeclareMathAlphabet{\mathsfit}{\encodingdefault}{\sfdefault}{m}{sl}
\SetMathAlphabet{\mathsfit}{bold}{\encodingdefault}{\sfdefault}{bx}{n}

\usepackage{hyperref}
\usepackage{url}
\usepackage{wrapfig}
\usepackage{graphicx}
\usepackage[ruled,vlined,noend]{algorithm2e}
\usepackage[normalem]{ulem}
\usepackage{float}
\usepackage{etoc}

\usepackage{xcolor}
\usepackage{enumitem}

\title{Pretrained Transformers As \\ Universal Computation Engines}

\author{Kevin Lu \\
UC Berkeley \\
\texttt{kzl@berkeley.edu} \\
\\
\textbf{Pieter Abbeel} \\
UC Berkeley \\
\texttt{pabbeel@cs.berkeley.edu}
\And
Aditya Grover \\
Facebook AI Research \\
\texttt{adityagrover@fb.com} \\
\\
\textbf{Igor Mordatch} \\
Google Brain \\
\texttt{imordatch@google.com}
}

\begin{document}

\maketitle

\begin{abstract}

We investigate the capability of a transformer pretrained on natural language to generalize to other modalities with minimal finetuning -- in particular, without finetuning of the self-attention and feedforward layers of the residual blocks.
We consider such a model, which we call a Frozen Pretrained Transformer (FPT), and study finetuning it on a variety of sequence classification tasks spanning numerical computation, vision, and protein fold prediction.
In contrast to prior works which investigate finetuning on the same modality as the pretraining dataset, we show that pretraining on natural language can improve performance and compute efficiency on non-language downstream tasks.
Additionally, we perform an analysis of the architecture, comparing the performance of a random initialized transformer to a random LSTM.
Combining the two insights, we find language-pretrained transformers can obtain strong performance on a variety of non-language tasks\footnote{
Code available at \href{https://github.com/kzl/universal-computation}{github.com/kzl/universal-computation}.
For a summary of changes made in the updated arXiv version, see Appendix \ref{app:changelog}.
}.

\end{abstract}

\begin{figure}[h]
    \centering
    \includegraphics[width=1\linewidth]{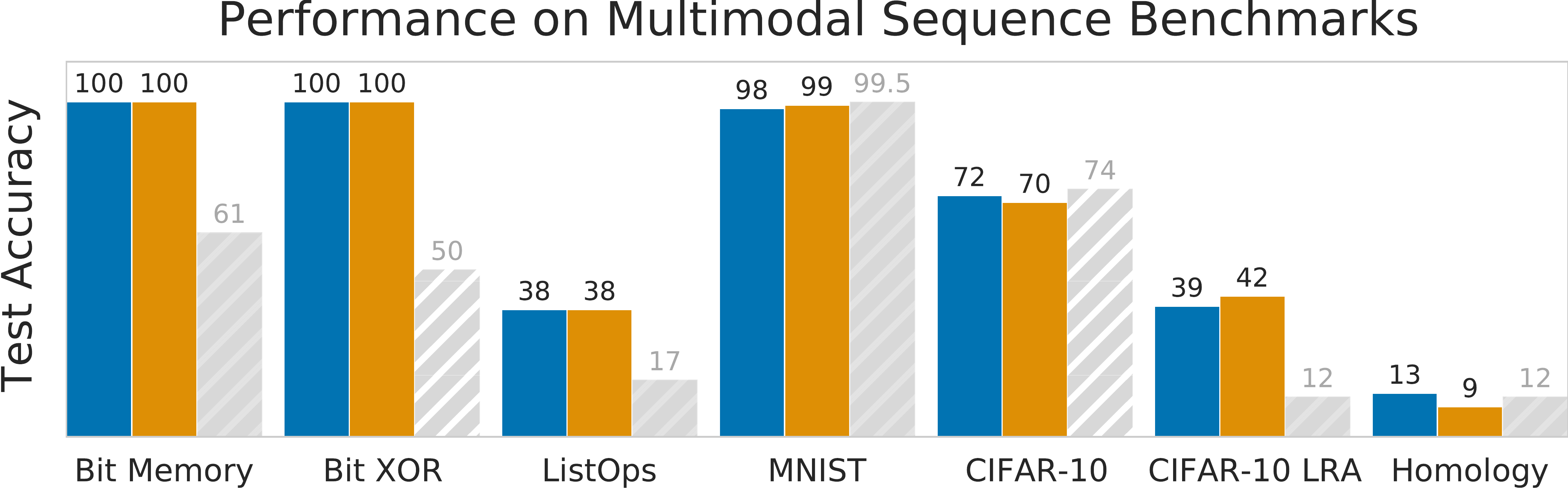}
    \vspace{0.5em}
    \includegraphics[width=0.9\linewidth]{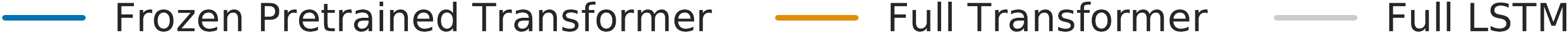}
    \caption{
    A \emph{frozen} language-pretrained transformer (FPT) -- without finetuning the self-attention and feedforward layers -- can achieve strong performance compared to a transformer fully trained from scratch on a downstream modality on benchmarks from literature \citep{tay2020lra, rap2019tape}.
    We show results on diverse classification tasks (see Section \ref{sec:tasks}): numerical computation (Bit Memory/XOR, ListOps), image classification (MNIST, CIFAR-10), and protein fold prediction (Homology).
    We also show results for a fully trained LSTM to provide a baseline.
    }
    \label{fig:main_result}
\end{figure}

\clearpage

\etocdepthtag.toc{mtchapter}
\etocsettagdepth{mtchapter}{subsection}
\etocsettagdepth{mtappendix}{none}
\tableofcontents

\clearpage

\section{Introduction}
\label{sec:intro}

The transformer architecture \citep{vaswani2017attention} has shown broad successes in deep learning, serving as the backbone of large models for tasks such as modeling natural language \citep{brown2020gpt3}, images \citep{dosovitskiy2020vit}, proteins \citep{jumper2021alphafold}, behaviors \citep{abramson2020imitating}, and multimodal tasks comprising of both images and text \citep{lu2019vilbert, radford2021clip}.
Inspired by these successes, we seek to explore the generalization capabilities of a transformer in transferring from one modality to another.

Classical approaches to sequence processing used recurrent neural network (RNN) approaches  \citep{rumelhart1985rnn, hochreiter1997lstm}.
In contrast, transformers utilize self-attention layers to extract features across tokens of a sequence, such as words \citep{vaswani2017attention} or image patches \citep{dosovitskiy2020vit}.
Furthermore, it has become common practice to train large models on unsupervised or weakly supervised objectives before finetuning or evaluating zero-shot generalization on a downstream task.
However, the downstream tasks that have been studied are generally restricted to the same modality as the original training set: for example, train GPT \citep{radford2018gpt} on a large language corpus, and finetune on a small task-specific dataset.
Our goal in this work is to investigate finetuning on modalities distinct from the training modality.

We hypothesize that transformers, namely the self-attention layers, can be pretrained on a data-rich modality (i.e. where data is plentiful, such as a natural language corpus) and identify feature representations that are useful for \emph{arbitrary} data sequences, enabling downstream transfer to different modalities.
In particular, we seek to investigate what pretrained language models (LMs) are capable of in terms of generalizing to other modalities with sequential structure.

To investigate this hypothesis, we take a transformer model pretrained on natural language data, GPT-2 \citep{radford2019gpt2}, and finetune only the linear input and output layers, as well as the positional embeddings and layer norm parameters.
We call this model a Frozen Pretrained Transformer (FPT).
On a range of tasks across a variety of modalities -- including numerical computation, image classification, and protein fold prediction -- FPT displays comparable performance to training the entire transformer or LSTM models from scratch, matching reported benchmarks for these tasks (Figure \ref{fig:main_result}).
Additionally, we find FPT models also converge faster during training.
Our results suggest the self-attention layers learned by a language model may have properties amenable to efficient universal computation.
Through a series of experiments, we seek to investigate what contributes to the performance of FPTs by isolating various sub-components of these models.

\section{Methodology}

\subsection{Tasks}
\label{sec:tasks}

We evaluate on a diverse set of classification tasks representative of different modalities.
In particular, we are interested in if language models are inherently capable of \emph{universal computation}, by which we mean the ability to learn representations for predictive learning across diverse modalities.

\textbf{Bit memory.}
Similar to the task proposed by \cite{miconi2018hebbian}, we consider a bit memory task where the model is shown 5 bitstrings each of length 1000.
Afterwards, the model is shown a masked version of one of the bitstrings, where each bit is masked with probability $0.5$, and the model is tasked with producing the original bitstring.
The bitstrings are broken up into sequences of length 50, so that the models are fed 120 tokens of dimension 50.

\textbf{Bit XOR.}
Similar to the bit memory task, the model is shown 2 bitstrings of length 5, where the model must predict the element-wise XOR of the two bitstrings.
The bitstrings are shown 1 bit at a time, so the models are fed 10 tokens of dimension 1.

\textbf{ListOps.}
Taken from \cite{tay2020lra}, the model is shown a sequence of list operations (ex. \texttt{[ MAX 4 3 [ MIN 2 3 ] 1 0 ]}) and tasked with predicting the resulting output digit (ex. \texttt{4}).
This task evaluates the ability of a model to parse mathematical expressions and evaluate over a long context.
The model is shown 1 token at a time, so the models are fed 512 tokens of dimension 15.

\textbf{MNIST.}
We use the standard MNIST benchmark, where the model must classify a handwritten digit from a $32 \times 32$ black-and-white image.
The tokens given to the model are $4 \times 4$ image patches, so the models are fed 64 tokens of dimension 16.

\textbf{CIFAR-10.}
We use the standard CIFAR-10 benchmark \citep{krizhevsky2009cifar}, where the tokens given to the model are $4 \times 4$ image patches, so the models are fed 64 tokens of dimension 16.

\textbf{CIFAR-10 LRA.}
This is a modified version of the above task taken from the Long Range Arena benchmark where the images are converted to grayscale and flattened with a token length of 1 \citep{tay2020lra}.
As a result, the input sequence consists of 1024 tokens of dimension 1.
This task is much more challenging than vanilla CIFAR-10 classification above as the models must learn patterns over a significantly longer sequence length and have minimal spatial inductive bias.

\textbf{Remote homology detection.}
In this task, we are interested in predicting the fold for a protein, represented as an amino acid sequence.
We use the datasets provided by TAPE \citep{rap2019tape, fox2013scop, hou2018deepsf}, where the train/test split is generated by holding out certain evolutionary groups.
Note that we do not pretrain on Pfam \citep{elgebali2019pfam}, which is common in other works.
There are 20 common and 5 uncommon amino acids (25 different types of inputs), and there are 1195 possible labels to predict.
We only consider sequences of length less than 1024 for simplicity.
The models are thus fed up to 1024 tokens of dimension 25.

\subsection{Architecture}
\label{sec:architecture}

\begin{figure}
    \centering
    \includegraphics[width=1\linewidth]{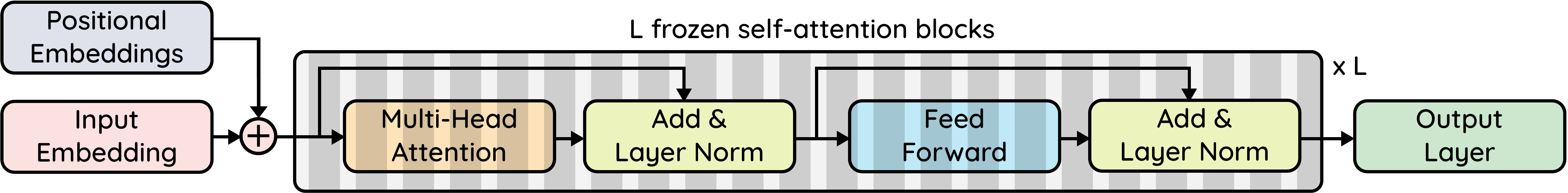}
    
    \caption{
        Frozen Pretrained Transformer (FPT).
        The self-attention \& feedforward layers are frozen.
    }
    \label{fig:architecture}
\end{figure}

The architecture we use is summarized in Figure \ref{fig:architecture}.
Denote the embedding size/hidden dimension of the transformer as $n_{dim}$, the number of layers as $n_{layers}$, (note $n_{dim} = 768$ and $n_{layers} = 12$ for the base size models), the input dimension as $d_{in}$, the output dimension (number of classes) as $d_{out}$, and the maximum length of the sequence as $l$.
We consider finetuning the following parameters of a pretrained GPT-2 model \citep{radford2019gpt2}:

\begin{itemize}[leftmargin=*]
    \item \textbf{Output layer:} it is crucial to finetune the output layer since we are transferring to a completely new task -- we use the simplest possible instantiation of an output network, being a single linear layer applied to the last output token output by the transformer, in order to highlight that almost all the computation is being performed by the frozen transformer.
    The output layer has $n_{dim} \times d_{out}$ parameters for the weight matrix.
    For example, for the base models on CIFAR-10, this comes out to $768 \cdot 10 = 7680$ parameters.

    \item \textbf{Input layer:} it is important to reinitialize a new input layer since we are reading in a new modality; in essence, we are learning how to query the transformer.
    This contrasts with prior unsupervised embedding evaluation techniques, such as linear probing -- due to the change in modality, we instead should train the input layer as well, and evaluate if the frozen intermediate transformer model performs effective computation.
    Again, we use a linear layer to minimize the amount of computation outside the transformer.
    The input layer has $d_{in} \times n_{dim}$ parameters for the weight matrix/embeddings, and an additional $n_{dim}$ parameters if there is a bias term.
    For the base models on CIFAR-10, this comes out to $16 \cdot 768 = 13056$ parameters.
    
    \item \textbf{Layer norm parameters:} as is standard practice in other finetuning works \citep{rebuffi2017adapter, houlsby2019adapter}, we also finetune the affine layer norm parameters (scale and bias), which adapt to the statistics of the downstream task in a new domain.
    In GPT-2, layer norm is applied twice per block, so these are a total of $4 \times n_{dim} \times n_{layers}$ parameters.
    For the base models on CIFAR-10, these come out to $4 \cdot 768 \cdot 12 = 36684$ parameters.
    
    \item \textbf{Positional embeddings:} While we observe that positional embeddings can be surprisingly universal between modalities (see Section \ref{sec:params}), we generally see a small benefit to finetuning the positional embeddings which have a cheap parameter cost of $l \times n_{dim}$.
    For the base models on CIFAR-10, these come out to $64 \cdot 768 = 49512$ parameters.
\end{itemize}
\vspace{2em}

Given the cheap linear scaling of these parameters, the parameter counts of large transformer models are dominated by the quadratic (in $n_{dim}$ and $l$) self-attention and feedforward layers.
For the base CIFAR-10 model with 124M parameters, these come out to approximately $0.086\%$ of the network.
Due to this scaling, this number decreases with larger model sizes, down to $0.029\%$ of the GPT-2 XL model.
We further ablate the importance of each parameter in Section \ref{sec:params}.
For more details and a description of the architecture, see Appendix \ref{app:architecture}.

Note that, crucially, all communication between tokens in the model are frozen.
The data in each datapoint is chunked into discrete tokens (bits, image patches, amino acids, etc.), and can only reference each other via the frozen attention connections, which are not trained; additionally, neither the output nor the input layers are connected to multiple tokens.
Our key investigation is to analyze the computation that is already inherent in the language model, and hence we do a minimal amount of computation that is learned on the downstream modality.

\section{Empirical Evaluations}
\label{sec:experiments}

In this section, we review the results demonstrating transfer from language to other modalities, and seek to better understand why this occurs and what enables this transfer.
All model sizes are the base model size (12 layers, 768 hidden dimension), unless stated otherwise.
See Appendix \ref{app:experimental_details} for more details on experiments.

\subsection{Can pretrained language models transfer to different modalities?}
\label{sec:transfer}

We investigate if the self-attention and feedforward layers -- the main body -- of a pretrained transformer can be applied to a classification problem in a different modality without finetuning.
To do this, we apply our base procedure as described above, where the input embedding layer, output readout layer, and layer norm parameters are finetuned.

Our results are shown in Figure \ref{fig:main_result} and also summarized below in Table \ref{table:main_result}.
We compare to state-of-the-art from literature when available (full transformer on ListOps, CIFAR-10 LRA, and Remote Homology; LSTM on Remote Homology).
Note the benchmarks from literature do not include decimal points, so for those numbers we report without a decimal.

We find that across all seven tasks considered, FPT achieves comparable performance to the fully trained transformer benchmarks.
We believe these results support the idea that these models are learning representations and performing computation that is agnostic to the modality.
We also note that both transformer variants significantly outperform LSTMs on some tasks, particularly ListOps and CIFAR-10 LRA, which have long sequence lengths of 512 and 1024, respectively.

On the two bit tasks (Memory and XOR), the models achieve 100\% performance, i.e. they are able to recover the exact algorithm.
Although our tables show results for $n=5$, we actually find FPT can still recover the exact algorithm on sequence lengths greater than $n=256$ (the elementwise XOR of two bitstrings each of length $256$), hinting that FPT has a fairly large working memory.

\begin{table}[h] 
\begin{center}
\begin{tabular}{c|ccccccc}
\toprule
\textbf{Model} & \multicolumn{1}{c}{\bf Bit Memory} & \multicolumn{1}{c}{\bf XOR} & \multicolumn{1}{c}{\bf ListOps} & \multicolumn{1}{c}{\bf MNIST} & \multicolumn{1}{c}{\bf CIFAR-10} & \multicolumn{1}{c}{\bf C10 LRA} & \multicolumn{1}{c}{\bf Homology} \\
\midrule
FPT & 100\% & 100\% & 38.4\% & 98.0\% & 72.1\% & 38.6\% & 12.7\% \\
Full & 100\% & 100\% & 38\% & 99.1\% & 70.3\% & 42\% & 9\% \\
LSTM & 60.9\% & 50.1\% & 17.1\% & 99.5\% & 73.6\% & 11.7\% & 12\% \\
\bottomrule
\end{tabular}
\end{center}
\caption{Test accuracy of FPT vs fully training transformer on downstream task vs fully training LSTM on downstream task (results are transcribed from Figure \ref{fig:main_result}).} \label{table:main_result}
\end{table}

We highlight a few important points for contextualizing these results.
We find that it can be difficult to fully train a 12-layer transformer on some of these (relatively small) datasets, as training can either diverge/overfit or be unstable.
For CIFAR-10, we report the full transformer results for a 3-layer model; for ListOps and CIFAR-10 LRA we report the number given for the 3-layer model from \cite{tay2020lra}; for Remote Homology we report the number for a smaller 12-layer model from \cite{rap2019tape}.
From an engineering perspective, this makes the full transformers harder to tune since we must choose model sizes that are stable and avoid overfitting -- see Section \ref{sec:generalization} for more analysis.
In particular, the numbers from \cite{tay2020lra} are generated from ``extensive sweeps over different hyper-parameters'' and use task-specific hyperparameters, while we do not tune the hyperparameters for FPT (except for remote homology; see Appendix \ref{app:experimental_details}).
In contrast, we find it is easy to improve the performance of FPT by increasing model size (see Section \ref{sec:size}) -- the CIFAR-10 number for FPT here is for the 36-layer large model.

Furthermore, unlike some other works utilizing transformers for vision, we use minimal spatial bias to emphasize the universal sequential aspect of the problem -- for instance, we do not interleave self-attention and convolution layers.
Note that we also do not use 2D positional embeddings (or other domain-specific techniques), hence providing very weak inductive prior to the model.
Our reasoning for these decisions is to evaluate the ability of transformers to work on arbitrary sequential tasks.

\subsection{What is the importance of the pretraining modality?}
\label{sec:pretraining}

We now compare pretraining on language to other pretraining methods for base model sizes:
\begin{itemize}[leftmargin=*]
    \item Random initialization (Random): initialization of the frozen transformer parameters randomly using the default initialization choices for GPT-2, i.e. without pretraining.
    
    \item Bit memory pretraining (Bit): pretraining from scratch on the Bit Memory task and then freezing the parameters before transferring.
    This allows the transformer to gain supervision working with arbitrary bit strings and performing memory/denoising on independent inputs.
    
    \item Image pretraining (ViT): using a pretrained Vision Transformer \citep{dosovitskiy2020vit} pretrained on ImageNet-21k \citep{deng2009imagenet}.
    Note that the architecture is a bit different, notably not using the autoregressive masking of GPT-2, since ViT is only pretrained on classification tasks (for other details, see Appendix \ref{app:details_pretraining}). 
\end{itemize}

These experiments highlight the significance of pretraining -- as opposed to simply the transformer architecture -- and compare language to other methods of supervision.
Our results are shown in Table \ref{table:random}.
Although the random transformers can achieve surprisingly strong accuracies, there is a considerable gap to using natural language pretraining, such as in MNIST, where random transformers achieve similar performance to a linear classifier on top of raw features (92\%).
Thus we believe that while the transformer architecture might be naturally conducive to these evaluations, the attention mechanisms used to transfer may be nontrivial and not fully specified by the architecture.
We also find that, in addition to performance benefits, language pretraining improves convergence compared to the randomly initialized transformer (see Section \ref{sec:compute_efficiency}).

\begin{table}[h] 
\begin{center}
\begin{tabular}{c|ccccccc}
\toprule
\textbf{Model} & \multicolumn{1}{c}{\bf Bit Memory} & \multicolumn{1}{c}{\bf XOR} & \multicolumn{1}{c}{\bf ListOps} & \multicolumn{1}{c}{\bf MNIST} & \multicolumn{1}{c}{\bf C10} & \multicolumn{1}{c}{\bf C10 LRA} & \multicolumn{1}{c}{\bf Homology} \\
\midrule
FPT & 100\% & 100\% & 38.4\% & 98.0\% & 68.2\% & 38.6\% & 12.7\% \\
Random & 75.8\% & 100\% & 34.3\% & 91.7\% & 61.7\% & 36.1\% & 9.3\% \\
Bit & 100\% & 100\% & 35.4\% & 97.8\% & 62.6\% & 36.7\% & 7.8\% \\
ViT & 100\% & 100\% & 37.4\% & 97.8\% & 72.5\% & 43.0\% & 7.5\% \\
\bottomrule
\end{tabular}
\end{center}
\caption{
Test accuracy of language-pretrained (FPT) vs randomly initialized (Random) vs Bit Memory pretraining (Bit) vs pretrained Vision Transformer (ViT) models.
The transformer is frozen.
}\label{table:random}
\end{table}

Pretraining on bit memory improves performance compared to the random models, but still lags behind training on natural language data.
Furthermore, measured by gradient steps, all models converge faster than the randomly initialized transformers (more details in Section \ref{sec:compute_efficiency}), indicating that all modes of pretraining improve upon random initialization even without considering accuracy.

Additionally, while freezing a vision transformer yields better improvements on CIFAR-10, pretraining on images is not uniformly better; e.g., ViT is worse on protein classification.
One hypothesis is that protein sequences are structured like language, in terms of discrete units of information with a ``grammar'', so transfer from language to proteins may be more natural.
\vspace{2em}

\subsection{How important is the transformer architecture compared to LSTM architecture?}
\label{sec:architecture_results}

In Section \ref{sec:pretraining} we found the transformer architecture can already be fairly effective in this regime, even with only random parameters.
In this section, we consider using a random LSTM architecture instead of the transformer, allowing us to consider the raw effect of architecture and ablating pretraining.
Like FPT, we finetune the input, output, and layernorm parameters for the LSTMs.

\begin{table}[h] 
\begin{center}
\begin{tabular}{c|ccccccc}
\toprule
\textbf{Model} & \multicolumn{1}{c}{\bf Bit Memory} & \multicolumn{1}{c}{\bf XOR} & \multicolumn{1}{c}{\bf ListOps} & \multicolumn{1}{c}{\bf MNIST} & \multicolumn{1}{c}{\bf CIFAR-10} & \multicolumn{1}{c}{\bf C10 LRA} & \multicolumn{1}{c}{\bf Homology} \\
\midrule
Trans.   & 75.8\% &  100\% & 34.3\% & 91.7\% & 61.7\% & 36.1\% & 9.3\% \\
LSTM     & 50.9\% & 50.0\% & 16.8\% & 70.9\% & 34.4\% & 10.4\% & 6.6\% \\
LSTM$^*$ & 75.0\% & 50.0\% & 16.7\% & 92.5\% & 43.5\% & 10.6\% & 8.6\% \\
\bottomrule
\end{tabular}
\end{center}
\caption{Test accuracy of randomly initialized transformers vs randomly initialized LSTM models. Note unlike in Figure \ref{fig:main_result}, the LSTM here is frozen. Frozen LSTMs perform very poorly. LSTM$^*$ represents an LSTM with additional architecture improvements to match the transformers (see below).}\label{table:random_architecture}
\end{table}
\vspace{-.5em}

Our results are shown in Table \ref{table:random_architecture}.
``LSTM'' refers to a 3-layer ``standard'' LSTM with a hidden dimension of 768, matching standard implementations of LSTMs, without residual connections or positional embeddings (see discussion below).
This matches the width of the FPT models, but not the depth or total parameter count (note that LSTMs also do not have positional embeddings).
We find that the self-attention architecture already serves as an effective inductive bias for universal computation, improving significantly over the recurrent LSTM model and comprising most of the improvement in test accuracy from random LSTM to FPT.

Here, we compare the 3-layer ``standard'' LSTM to a 12-layer ``standard'' LSTM.
Note that most LSTM implementations, including the one used in Table \ref{table:random_architecture}, do not feature residual connections and positional embeddings.
We include this comparison to represent the traditional method more faithfully, but add these additional architectural components below.
In the same style of FPT and GPT-2, we do not use a bidirectional LSTM.
Under these model choices, we report the performance of a frozen random 3-layer vs 12-layer LSTM in Table \ref{table:lstm_layers}.
Naively, the 12-layer model is much worse than the 3-layer model, hinting that there is some loss of information by repeated LSTM layers.

\begin{table}[h] 
\begin{center}
\begin{tabular}{c|cccc}
\toprule
\textbf{Layers} & \multicolumn{1}{c}{\bf ListOps} & \multicolumn{1}{c}{\bf MNIST} & \multicolumn{1}{c}{\bf CIFAR-10} & \multicolumn{1}{c}{\bf C10 LRA} \\
\midrule
12 & 16.2\% & 11.7\% & 10.8\% & 10.4\% \\
3  & 16.8\% & 70.9\% & 34.4\% & 10.4\% \\
\bottomrule
\end{tabular}
\end{center}
\caption{Test accuracy of randomly initialized ``standard'' LSTMs varying number of layers with a hidden dimension of 768. The simple 12-layer LSTM achieves only near-trivial performance.}\label{table:lstm_layers}
\end{table}
\vspace{-.5em}

We also experiment with ablating other architectural improvements included with the transformer architecture in Table \ref{table:lstm_layers_residual}.
Once residual connections \citep{he2016resnet} are added, the 12-layer LSTM makes up a lot of the performance drops, hinting that residual connections could make up for loss of information from the LSTM layers which otherwise linearly combine the features.
We also add positional embeddings, which finishes bridging the gap between standard LSTM implementations and the transformer.
Even with these additional benefits, the LSTM still performs worse.
Note that the final 12-layer LSTM has about the same number of trainable parameters as the transformer.

\begin{table}[h] 
\begin{center}
\begin{tabular}{c|cccc}
\toprule
\textbf{Model} & \multicolumn{1}{c}{\bf ListOps} & \multicolumn{1}{c}{\bf MNIST} & \multicolumn{1}{c}{\bf CIFAR-10} & \multicolumn{1}{c}{\bf C10 LRA} \\
\midrule
12-Layer LSTM           & 16.2\% & 11.7\% & 10.8\% & 10.4\% \\
+ Residual Connections  & 16.8\% & 70.9\% & 34.4\% & 10.4\% \\
+ Positional Embeddings & 16.7\% & 92.5\% & 43.5\% & 10.6\% \\
\midrule
Random Transformer      & 34.3\% & 91.7\% & 61.7\% & 36.1\% \\
\bottomrule
\end{tabular}
\end{center}
\caption{Test accuracy of 12-layer randomly initialized ``standard'' LSTMs additional architectures modifications to match transformers: residual connections and positional embeddings.
The bottom row, LSTM with residual connections and positional embeddings, is nearly identical to GPT-2.}\label{table:lstm_layers_residual}
\end{table}

\subsection{Does language pretraining improve compute efficiency over random initialization?}
\label{sec:compute_efficiency}

We investigate compute efficiency by considering the number of gradient steps to converge for FPT vs random transformer models, shown in Table \ref{table:convergence}.
We generally find FPT converges faster, which indicates language pretrainining can yield compute benefits for non-language tasks.
While random transformer models achieve decent test accuracies, in particular when compared to random LSTMs, there is still a considerable gap in the compute efficiency compared to using pretraining.
Note that bit memory pretraining introduced in Section \ref{sec:pretraining} generally falls between the two models, and notably is $6 \times$ slower than FPT on Bit XOR, which is significantly better than random.

\begin{table}[h] 
\begin{center}
\begin{tabular}{c|ccccccc}
\toprule
\textbf{Model} & \multicolumn{1}{c}{\bf Memory} & \multicolumn{1}{c}{\bf XOR} & \multicolumn{1}{c}{\bf ListOps} & \multicolumn{1}{c}{\bf MNIST} & \multicolumn{1}{c}{\bf C10} & \multicolumn{1}{c}{\bf C10 LRA} & \multicolumn{1}{c}{\bf Homology} \\
\midrule
FPT & $1 \times 10^4$ & $5 \times 10^2$ & $2 \times 10^3$ & $5 \times 10^3$ & $4 \times 10^5$ & $3 \times 10^5$ & $1 \times 10^5$\\
Random & $4 \times 10^4$ & $2 \times 10^4$ & $6 \times 10^3$ & $2 \times 10^4$ & $4 \times 10^5$ & $6 \times 10^5$ & $1 \times 10^5$ \\
\midrule
\textbf{Speedup} & $4 \times$ & $40\times$ & $3 \times$ & $4 \times$ & $1 \times$ & $2 \times$ & $1 \times$ \\
\bottomrule
\end{tabular}
\end{center}
\caption{Approximate number of gradient steps until convergence for pretrained (FPT) vs randomly initialized (Random) models. Note that we use the same batch size and learning rate for both models.}\label{table:convergence}
\end{table}

\subsection{Do the frozen attention layers attend to modality-specific tokens?}
\label{sec:attention_maps}

\begin{figure}[H]
    \centering
    \includegraphics[width=0.55\linewidth]{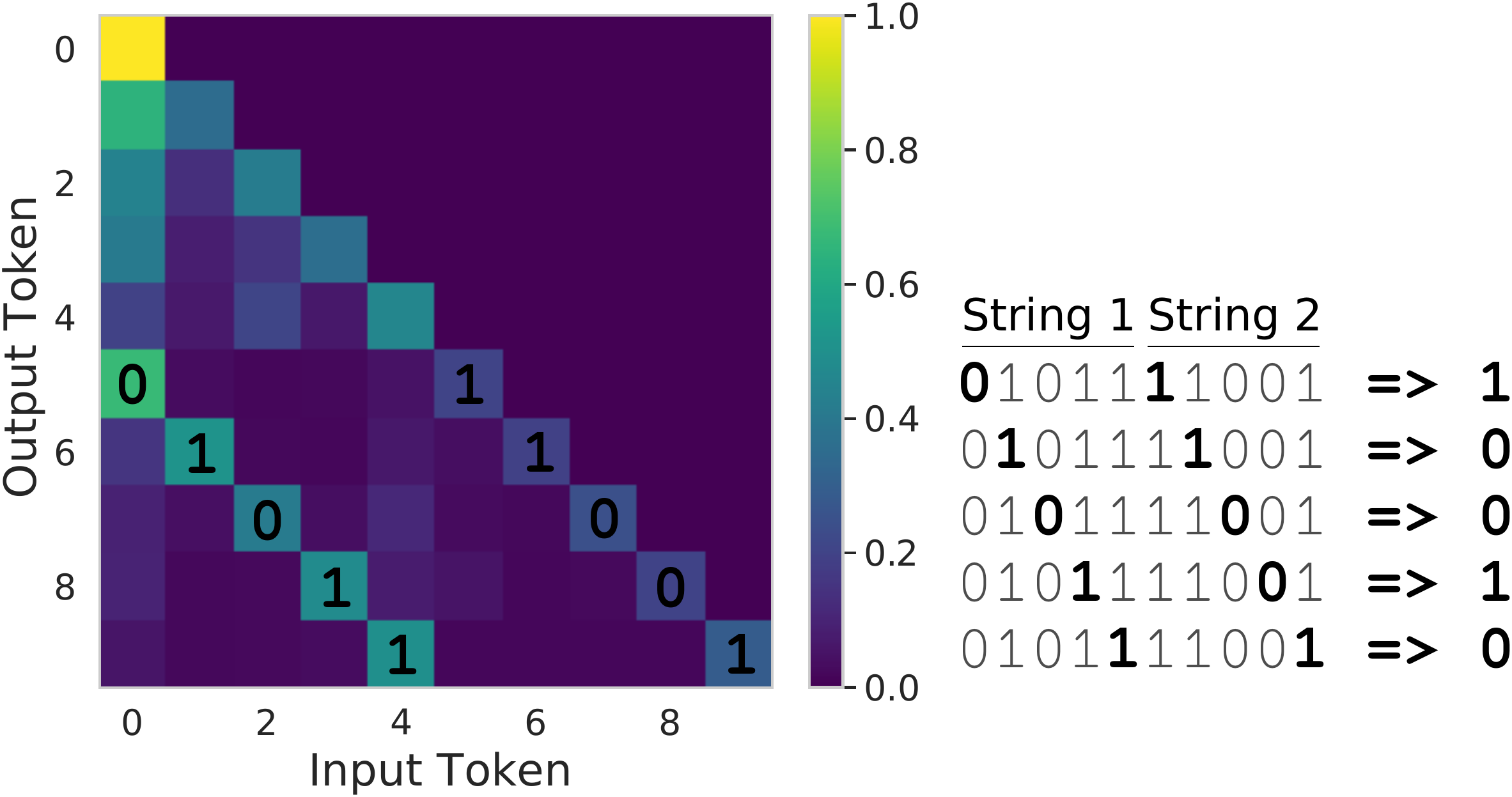}
    \caption{
        On Bit XOR, the model must produce the element-wise XOR of two bitstrings presented sequentially (inputs 0-4 are the first bitstring, inputs 5-9 are the second).
        Each token is one bit.
        FPT learns to attend positionally to the two bits that are XOR'ed by the output token.
    }
    \label{fig:attn_xor_pretrained}
\end{figure}

\begin{figure}[H]
    \centering
    \includegraphics[width=.95\linewidth]{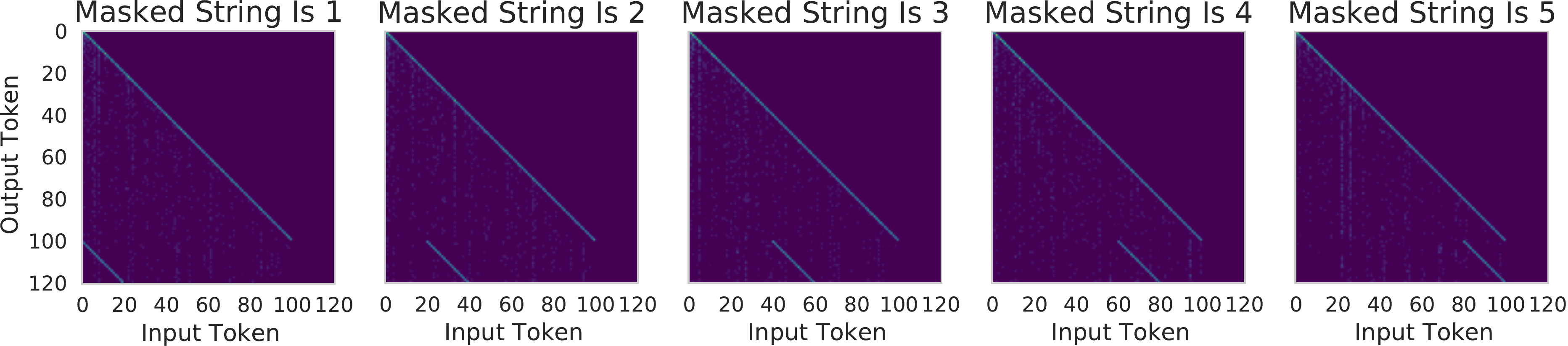}
    \caption{
        On Bit Memory, the model must return one of five strings (inputs 0-99) given a masked version of one of the strings (inputs 100-119).
        Each token is 50 bits.
        FPT learns to attend to the correct string based on finding similarity to the inputs, not relying solely on position as in Bit XOR.
    }
    \label{fig:attn_memory_pretrained}
\end{figure}

We investigate if FPT attends to semantically meaningful patterns in the data.
We plot the attention weights (i.e. the values of the softmax of query-key dot product) from the first layer.
We show the results in Figures \ref{fig:attn_xor_pretrained} and \ref{fig:attn_memory_pretrained} for the bit tasks.
Note GPT-2 is autoregressive, so the upper right corner of the attention mask is zeroed out.
On these tasks, FPT yields an interpretable attention pattern despite not training the self-attention layers themselves.
We did not find easily interpretable patterns on the other tasks.

\begin{figure}[H]
    \centering
    \includegraphics[width=0.3\linewidth]{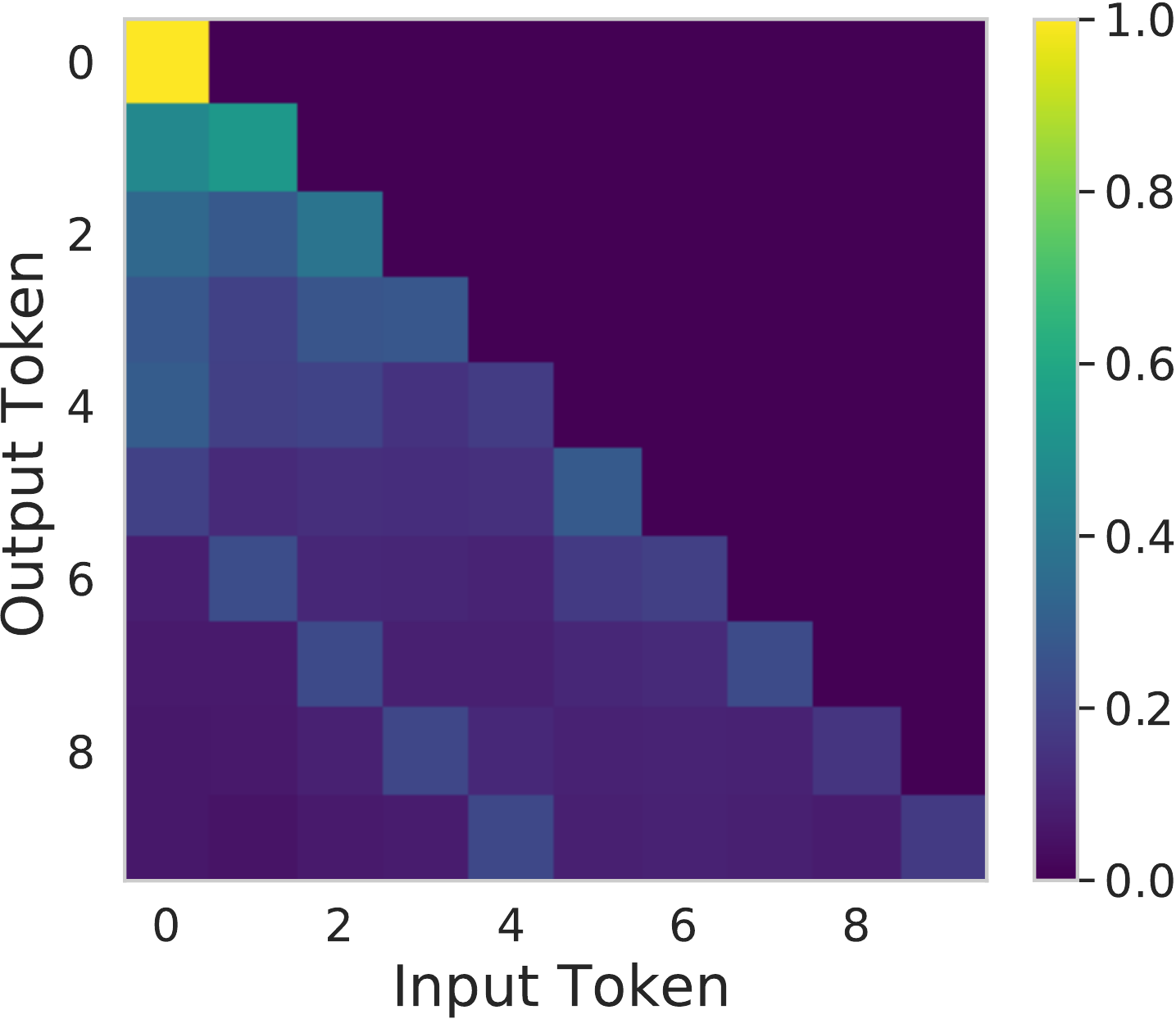}
    \caption{
        A transformer with frozen randomly initialized self-attention layers also learns to correlate the two diagonal elements on Bit XOR, although the magnitude of the diagonals is lower (note the extra attention weights distributed in between the diagonals).
    }
    \label{fig:attn_xor_random}
\end{figure}

We also include the attention map for Bit XOR using a randomly initialized transformer (which also solves the task) in Figure \ref{fig:attn_xor_random}.
This model also learns to exploit the diagonal pattern, although the strength is a little weaker.
This indicates that while the random transformer still learns to solve the task, it learns a less semantically interpretable/strong attention pattern.

\subsection{Does freezing the transformer prevent overfitting or underfitting?}
\label{sec:generalization}

Our general findings are that -- in contrast to their fully trained counterparts -- FPT models underfit the data, which lends them to further improvements by increasing model capacity (see Section \ref{sec:size}).
For example, consider CIFAR-10 LRA, which is maximally difficult due to lack of inductive prior over the sequence (each pixel is fed in as an arbitrary token only ordered by a raster scan) and relatively small dataset (50k images).
In Table \ref{table:generalization}, we show the train/test gap between training FPT vs a 3-layer transformer from \cite{tay2020lra}, which we find to give stronger results than our experiments.
In particular, they are much better than training a 12-layer transformer, which works poorly.
Our results indicate that FPT is generally providing generalizable task representations without causing overfitting, whereas transformers can overfit arbitrarily poorly in low-data regimes (such as for Linformer, which overfit the most out of the architectures tested by \cite{tay2020lra}).
More work can investigate how to increase the model expressiveness, which could yield performance benefits.

\begin{table}[h] 
\begin{center}
\begin{tabular}{c|c|cc}
\toprule
\textbf{Model} & \textbf{\# Layers} & \multicolumn{1}{c}{\bf Test Accuracy} & \multicolumn{1}{c}{\bf Train Accuracy} \\
\midrule
FPT (GPT-2) & 12 & 38.6\% & 38.5\% \\
Vanilla Transformer & 3 & 42\% & 70\% \\
Linformer & 3 & 39\% & 97\% \\
\bottomrule
\end{tabular}
\end{center}
\caption{Train vs test accuracies on CIFAR-10 LRA task.}\label{table:generalization}
\end{table}



\subsection{Does performance scale with model size?}
\label{sec:size}

We evaluate the efficacy of adding more parameters to these models on CIFAR-10.
Most of the additional parameters are in the transformer layers and are trained during the natural language pretraining phase.
Our results for pretrained and random models are in Table \ref{table:larger_models}.
Unlike fully training a transformer, which exhibits more overfitting and divergence during training with larger models, increasing model size stably increases the capacity of the models.
This result indicates our observations and results are likely to scale as we move towards larger models and higher-data regimes.

\begin{table}[h] 
\begin{center}
\begin{tabular}{c|ccc|cc}
\toprule
\textbf{Model Size} & \multicolumn{1}{c}{\bf \# Layers} & \multicolumn{1}{c}{\bf Total Params} & \textbf{Trained Params} & \multicolumn{1}{c}{\bf FPT} & \multicolumn{1}{c}{\bf Random} \\
\midrule
Small (Base) & 12 & 117M & 106K & 68.2\% & 61.7\% \\
Medium       & 24 & 345M & 190K & 69.8\% & 64.0\% \\
Large        & 36 & 774M & 300K & 72.1\% & 65.7\% \\
\bottomrule
\end{tabular}
\end{center}
\caption{Test accuracy of larger frozen transformer models on CIFAR-10.}\label{table:larger_models}
\end{table}

\subsection{Can performance be attributed simply to better statistics for initialization?}
\label{sec:initialization}

\begin{figure}[H]
    \centering
    \includegraphics[width=1\linewidth]{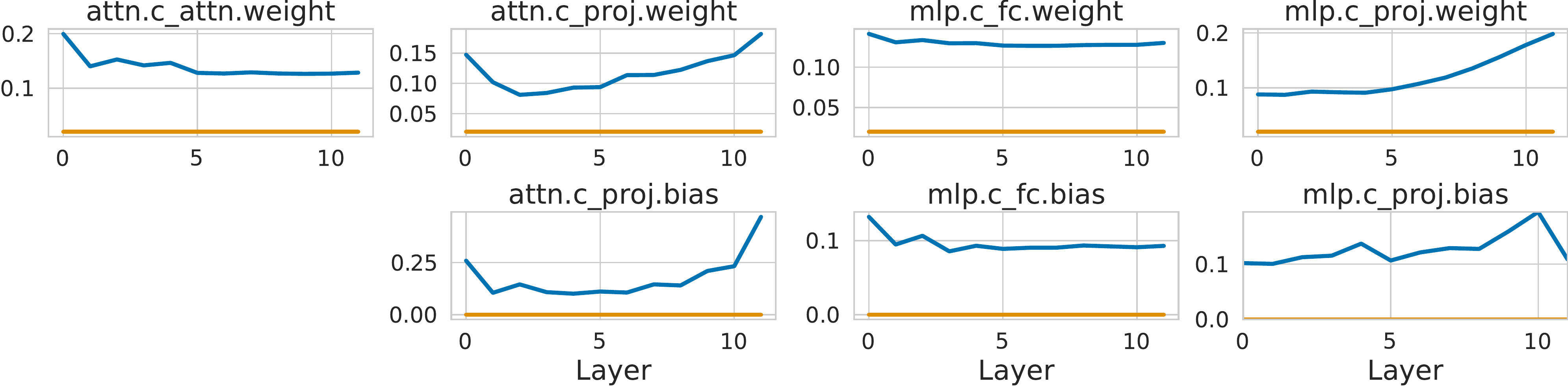}
    \vspace{1em}
    \includegraphics[width=0.6\linewidth]{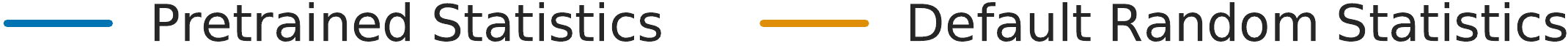}
    \caption{
        Standard deviation of the parameters by layer for the pretrained GPT-2 model versus default initialization hyperparameters ($0.02$ for weights and $0$ for biases).
    }
    \label{fig:statistics}
\end{figure}

In this section, we ablate taking the layer-wise mean and standard deviation from the pretrained model and using it to initialize a random transformer, in order to ablate if a better initialization scheme via an ``oracle'' standard deviation can recover the performance of FPT.
Note that the GPT-2 initialization scheme initializes parameters as Gaussian; traditionally, the standard deviation is $0.02$ by default.
For clarity, we show the standard deviation by layer for the weights and biases of the attention and feedforward layers in Figure \ref{fig:statistics} for the pretrained models.

\vspace{-1em}

We show the results using this initialization scheme in Table \ref{table:initialization} (note that all of the weights, biases, layer norm, and positional embeddings are initialized -- both mean and variance -- in this fashion).
This yields better results on most tasks, but does poorly on CIFAR-10.
As a result, we believe the benefits of language pretraining cannot be recovered with a simple better initialization scheme, although we believe future work in transformer initialization could yield different results.

\begin{table}[H] 
\begin{center}
\begin{tabular}{c|ccccccc}
\toprule
\textbf{Initialization} & \multicolumn{1}{c}{\bf Memory} & \multicolumn{1}{c}{\bf XOR} & \multicolumn{1}{c}{\bf ListOps} & \multicolumn{1}{c}{\bf MNIST} & \multicolumn{1}{c}{\bf C10} & \multicolumn{1}{c}{\bf C10 LRA} & \multicolumn{1}{c}{\bf Homology} \\
\midrule
Pretrained & 100\% & 100\% & 38.4\% & 98.0\% & 68.2\% & 38.6\% & 12.7\% \\
Statistics Only & 100\% & 100\% & 37.4\% & 97.2\% & 56.5\% & 33.1\% & 11.0\% \\
Default & 75.8\% & 100\% & 34.3\% & 91.7\% & 61.7\% & 36.1\% & 9.3\% \\
\bottomrule
\end{tabular}
\end{center}
\caption{Test accuracy when initializing parameters with pretrained weights (i.e., FPT) vs randomly initializing parameters according to the mean and variance of the pretrained transformer (Statistics Only) vs random initialization with default parameters (Default).}\label{table:initialization}
\end{table}
\vspace{-1.8em}

\subsection{Can we train a transformer by only finetuning the output layer?}
\label{sec:reservoir}

We consider using FPT solely for naive feature extraction for linear classification, where we fix a randomly initialized input layer and freeze all parts of the model except for the output.
Note that this resembles resevoir computing/echo state networks (see Section \ref{sec:gwt} for discussion).
The model evaluates on every example in the training set once, caches the features, and then we train a linear output layer.
This enables subsequent epochs after the first to run extremely quickly, but does not easily handle dropout/data augmentations, and scales well in terms of number of epochs, but not in dataset size.
Note that this is mathematically equivalent to linear classification.
Our results are shown in Table \ref{table:linear}.
Although we find speedups extremely significant and they obtain nontrivial performance, performance significantly degrades and the models also exhibit overfitting (likely due to lack of regularization; unlike the training of FPT, dropout is not applied).

\begin{table}[H]
\begin{center}
\begin{tabular}{c|c|ccc}
\toprule
\textbf{Task} & \textbf{Speedup} & \multicolumn{1}{c}{\bf Output Only} & \multicolumn{1}{c}{\bf FPT} & \multicolumn{1}{c}{\bf Full Transformer}  \\
\midrule
ListOps      & $500-2000\times$ & 32.8\% & 38.4\% & 38\% \\
CIFAR-10 LRA & $500-2000\times$ & 24.7\% & 38.6\% & 42\% \\
\bottomrule
\end{tabular}
\end{center}
\caption{Training only the output layer as a linear regression problem. Speedup refers to wall clock time per epoch (after the first). Larger models have larger speedups.}\label{table:linear}
\end{table}

\subsection{What is the role of model depth in token mixing?}
\label{sec:model_depth}

One interesting question is the importance of the depth of the transformer for generating representions which ``mix'' tokens: for instance, if there is only one layer and the parameters are random, it is unlikely for the tokens to be mixed well, whereas if there are many layers, there are many chances for the tokens to mix and form interesting representations useful for downstream tasks.
We investigate this on ListOps by considering pretrained vs random models, where we only take the first X layers of the 12-layer pretrained model (i.e. for X=3, we use the first 3 layers of the pretrained GPT-2 model and perform classification from those hidden states).
Additionally, to maximally highlight the importance of the pretrained parameters, we randomly initialize the input layer, and do not train the input or positional parameters.
We first show results are finetuning the output layer and layernorm parameters, and then show only finetuning the output layer.

\textbf{With finetuning layernorm.}
We first investigate this question with finetuning the layernorm parameters (i.e. we finetune only the output layer and the layernorm parameters).
Results are shown in Table \ref{table:depth_ln}.
Both models are unable to do well with only one layer, but the pretrained model performs significantly better than the random model at 2 layers, indicating that while the difference in performance at 12 layers is relatively small, there is a great benefit to using pretrained layers for when considering a small number of layers in that the tokens are ``mixed'' faster.

\begin{table}[h] 
\begin{center}
\begin{tabular}{c|cc}
\toprule
\textbf{Number of Layers} & \multicolumn{1}{c}{\bf Pretrained} & \multicolumn{1}{c}{\bf Random} \\
\midrule
1 & 17\% & 17\% \\
2 & 36\% & 16\% \\
6 & 38\% & 35\% \\
\bottomrule
\end{tabular}
\end{center}
\caption{Test accuracy on Listops while varying model depth and finetuning layernorm parameters. Pretrained layers ``mix'' the tokens faster, performing better at low model depths.}\label{table:depth_ln}
\end{table}

\textbf{Without finetuning layernorm.}
We now investigate this question without finetuning the layernorm parameters, and only finetuning the output parameters, as in the reservoir computing setup in Section \ref{sec:reservoir}.
Note this is equivalent to linear classification.
This setting is the most challenging since all processing that is able to mix tokens is done by either random or pretrained parameters, and we are only able to train a linear layer on top of the output of the last token; as a result, the \emph{only} token mixing that is done is performed entirely by the pretrained self-attention layers.
Results are shown in Table \ref{table:depth_no_ln}.
The random model does not do well even for a large number of layers, while the pretrained model can still do reasonably well, even though it requires more layers than before.

\begin{table}[h] 
\begin{center}
\begin{tabular}{c|cc}
\toprule
\textbf{Number of Layers} & \multicolumn{1}{c}{\bf Pretrained} & \multicolumn{1}{c}{\bf Random} \\
\midrule
1 & 12\% & - \\
3 & 18\% & - \\
6 & 33\% & - \\
12 & 33\% & 17\% \\
24 & - & 17\% \\
\bottomrule
\end{tabular}
\end{center}
\caption{Test accuracy on Listops while varying model depth and only training output parameters. Even for a large number of layers, the random model does not learn to perform well.}\label{table:depth_no_ln}
\end{table}

\subsection{Can training more parameters improve performance?}
\label{sec:moreparams}

Our focus in this work was primarily to investigate if and how efficient, general-purpose pretraining can transfer across modalities.
However, for practical applications, it would naturally be more suited to choose a more specialized finetuning scheme or add more trainable parameters.
In this section, we investigate additionally finetuning parameters with various methods, to see if frozen language transformers can serve as a practical base for future work.

We first investigate additionally finetuning the self-attention and feedforward layers, which were previously frozen.
We simply add them to the list of parameters finetuned, without changing the optimization or learning rate scheme, although this is suboptimal.
Our results are shown in Table \ref{table:finetune_attn_ff}.
Note that +Both is fully finetuning the 12-layer transformer (in other sections, we use full transformer to denote fully finetuning a transformer from scratch where the depth was tuned, whereas here the depth is fixed).
We find that finetuning the feedforward layers can improve performance, which is similar to techniques used in prior work \citep{houlsby2019adapter}, but finetuning the attention layers can lead to divergence.

\begin{table}[H] 
\begin{center}
\begin{tabular}{c|ccccccc}
\toprule
\textbf{Model} & \multicolumn{1}{c}{\bf Memory} & \multicolumn{1}{c}{\bf XOR} & \multicolumn{1}{c}{\bf ListOps} & \multicolumn{1}{c}{\bf MNIST} & \multicolumn{1}{c}{\bf C10} & \multicolumn{1}{c}{\bf C10 LRA} & \multicolumn{1}{c}{\bf Homology} \\
\midrule
FPT           & 100\% & 100\% & 38.4\% & 98.0\% & 68.2\% & 38.6\% & 12.7\% \\
+ Feedforward & 100\% & 100\% & 36.0\% & 98.3\% & 76.6\% & 38.2\% & 13.1\% \\
+ Attention   & 100\% & 100\% & 36.8\% & 89.0\%$^\dagger$ & 47.7\%$^\dagger$ & 23.0\% & 10.9\% \\
+ Both        & 100\% & 100\% & 35.8\% & 93.1\%$^\dagger$ & 32.9\% & 21.0\% & 10.5\% \\
\bottomrule
\end{tabular}
\end{center}
\caption{
    Additionally finetuning either the feedforward layers, attention layers, or both.
    We do not use a per-layer learning scheme/etc.
    $^\dagger$training diverged, number reported before divergence.
}\label{table:finetune_attn_ff}
\end{table}

On CIFAR-10, we experiment with additionally finetuning the last attention layer, shown in Table \ref{table:morelayers}.
Generally we find smarter pretraining methods can yield better performance, so we are optimistic about the possibility of multimodal training/architectures \emph{improving} performance in future work.

\begin{table}[H] 
\begin{center}
\begin{tabular}{cccc}
\toprule
\textbf{Task} & \multicolumn{1}{c}{\bf Base (FPT)} & \multicolumn{1}{c}{\bf + Finetuning All FF Layers} & \multicolumn{1}{c}{\bf + Finetuning Last Attn Layer} \\
\midrule
CIFAR-10 & 68.2\% & 76.6\% & 80.0\% \\
\bottomrule
\end{tabular}
\end{center}
\caption{Test accuracy on CIFAR-10 when finetuning additional parameters. In addition to FPT, if we finetune the feedforward layers and the last self-attention layer, we can achieve 80\% accuracy. }\label{table:morelayers}
\end{table}

\subsection{Which parameters of the model are important to finetune?}
\label{sec:params}

We now run ablations for only finetuning select parameters to see which parameters are most sensitive.
Note for all experiments (including the previous ones), we initialize the input layers as Gaussian if embeddings are used, or use an orthogonal initialization for linear layers; in particular, we find orthogonal initialization to be very important when input parameters are not trained.
We highlight some results in Table \ref{table:finetuning_add}; full results are shown on Page~\pageref{table:finetuning_indep}.
Similar to a study of random CNNs by \cite{frankle2020batchnorm}, we generally find the layer norm parameters to be most important.

\begin{table}[H]
\begin{center}
\begin{tabular}{c|cccc}
\toprule
\textbf{Task} & \multicolumn{1}{c}{\bf output only} & \multicolumn{1}{c}{\bf + layernorm} & \multicolumn{1}{c}{\bf + input} & \multicolumn{1}{c}{\bf + positions} \\
\midrule
Bit Memory & 76\% & 94\% & 100\% & 100\% \\
Bit XOR & 56\% & 98\% & 98\% & 100\% \\
ListOps & 15\% & 36\% & 36\% & 38\% \\
MNIST & 23\% & 96\% & 98\% & 98\% \\
CIFAR-10 & 25\% & 54\% & 60\% & 68\% \\
CIFAR-10 LRA & 17\% & 39\% & 39\% & 39\% \\
Homology & 2\% & 9\% & 10\% & 13\% \\
\bottomrule
\end{tabular}
\end{center}
\caption{Ablation by successively adding certain parameters to the list of finetuned parameters for pretrained frozen transformers.}\label{table:finetuning_add}
\end{table}

\subsection{Is finetuning layer norm necessary for FPT to perform well?}

While previously we showed performance gains with finetuning layer norm, we could instead consider only finetuning the input and output layers, treating the entire GPT model as a black box.
We show results on CIFAR-10 in Table \ref{table:nolayernorm}.
The model performs worse; note accuracy is similar to not finetuning the positional embeddings (see Section \ref{sec:params}).
This suggests the internal modulation of the affine layer norm parameters help, possibly by about as much as finer positional information.

\begin{table}[H] 
\begin{center}
\begin{tabular}{ccc}
\toprule
\textbf{Initialization} & \multicolumn{1}{c}{\bf Frozen Layer Norm} & \multicolumn{1}{c}{\bf Finetuned Layer Norm} \\
\midrule
Pretrained & 61.5\% & 68.2\%\\
Random     & 55.0\% & 61.7\% \\
\bottomrule
\end{tabular}
\end{center}
\caption{Test accuracy on CIFAR-10 when only finetuning the input and output layer parameters.}\label{table:nolayernorm}
\end{table}

\subsection{How well do the trends hold across other transformer models?}
\label{sec:alternative_architectures}

We also investigate how other transformer architectures perform when swapped out with GPT-2: BERT \citep{devlin2019bert}, T5 \citep{raffel2019t5}, and Longformer \citep{beltagy2020longformer}.
For T5, we only use the encoder, and not the decoder.
Our results are in Table \ref{table:nlp_architectures}.
We find results to roughly hold across some architectures -- with some differences -- although T5 tends to be slightly worse than the other models.
An interesting question for future work is whether subtle differences in architecture, pretraining objective, or dataset contribute to these differences.

\begin{table}[h] 
\begin{center}
\begin{tabular}{c|cccc}
\toprule
\textbf{Task} & \multicolumn{1}{c}{\bf GPT-2 (FPT Default)} & \multicolumn{1}{c}{\bf BERT} & \multicolumn{1}{c}{\bf T5} & \multicolumn{1}{c}{\bf Longformer} \\
\midrule
ListOps & 38.4\% & 38.3\% & 15.4\% &  17.0\% \\
CIFAR-10 & 68.2\% & 68.8\% & 64.7\% & 66.8\% \\
\bottomrule
\end{tabular}
\end{center}
\caption{Test accuracy for frozen pretrained transformer variants (base model sizes).}\label{table:nlp_architectures}
\end{table}

\section{Related Work and Discussion}
\label{sec:relatedwork}

\subsection{Transformers in multimodal settings}

Transformers \citep{vaswani2017attention} were first used successfully for natural language processing \citep{radford2018gpt, devlin2019bert, radford2019gpt2, brown2020gpt3}.
In recent years, they have also been shown to be effective architectures for other modalities.
One particular modality of interest is computer vision \citep{chen2020imagegpt, touvron2020deit}; in particular, \cite{dosovitskiy2020vit} showed that transformers can outperform CNNs in the high-data regime on standard object recognition benchmarks such as ImageNet and CIFAR.
Furthermore, transformers have also been used for prediction tasks over protein sequences~\citep{jumper2021alphafold, rao2021msa}, reinforcement learning \citep{parisotto2020stabilizing}, and imitation learning \citep{abramson2020imitating}.

Work specifically tackling multimodal tasks include \cite{kaiser2017multitask}, who showed a single model could learn a variety of multimodal tasks with an attention architecture.
Recent work has utilized transformers for multimodal predictive tasks, such as images and text in ViLBERT \citep{lu2019vilbert} and CLIP \citep{radford2021clip}; these approaches generally use two distinct transformers to embed images and text.
\cite{lu2020vilbertmulti} applies ViLBERT to train a single model for a variety of combined vision and language tasks.
Recent work from OpenAI \citep{goh2021multimodal} finds that some neurons learned by CLIP are activated by a particular semantic concept, regardless of if the concept is presented in language or picture form.
Our work is most similar to DALL-E \citep{ramesh2021dalle}, which uses a single transformer to embed both the image and text modalities, which we consider to be generating a ``universal latent space'' that projects any type of input into a single latent space.
Such a latent space would be useful for a model that could learn from many sources of supervision.

\subsection{Transformers in transfer settings}

There are also many works looking at transformers specifically in the context of in-modality transfer, such as ViT for vision \citep{dosovitskiy2020vit}, T5 for language \citep{raffel2019t5}, and UDSMProt for protein sequences \citep{strodthoff2020udsmprot}.
CLIP~\citep{radford2021clip} showed that training on text in addition to images could allow for zero-shot classification via providing downstream labels as text.
\cite{hernandez2021scaling} do a thorough investigation of transfer with language pretraining, notably showing transfer from English to Python, which they consider to be reasonably distanced from English; many works have also looked at transferring from one langauge to another \citep{artetxe2019cross, ponti2019towards}.
Similar to our work, \cite{papadimitriou2020music} investigate transfer for LSTMs between modalities including code, different languages, and music, finding that pretraining on ``non-linguistic data with latent structure'' can transfer to language, finding grammatical structure in a modality to be important, although we generally investigate the other direction and explore more distanced modalities. 
\cite{kiela2019supervised} make similar observations for aligning representation spaces of language and vision.
\cite{li2020communication} pretrain on a referential communication game where an emergent learned language is used to transfer to NLP tasks.
\cite{wu2021lime} found explicitly pretraining computational primitives to transfer to mathematics tasks.

\subsection{Pretraining and finetuning of transformer models}

A common trend in deep learning models is to first train a large model on an unsupervised objective on a large dataset \citep{dai2015semi, radford2018gpt} and then finetune on a small downstream dataset (e.g., by freezing the model and only finetuing the output layer).
A common method used to finetune transformers are adapter networks \citep{rebuffi2017adapter, houlsby2019adapter}, which add a fully connected residual block for each unique downstream task and also finetune the layer norm parameters.
For simplicity, we do not add the full adapter block but only train the layer norm parameters, reducing the number of parameters we consider.
These techniques used are similar to prior approaches such as FiLM \citep{perez2018film} and self-modulation \citep{chen2018selfmodulation}.
A recent direction of research has explored learning prompt templates for large models \citep{shin2020autoprompt} that simply require forward passes over the transformer.
Unlike these works, we consider finetuning on one modality (language) and finetuning on others, whereas prior work investigates finetuning on the same modality as the pretraining task.
Another interesting related work, although not investigating transformers, by \cite{frankle2020batchnorm} find randomly initialized CNNs, which only train the batchnorm affine parameters, to work well on CIFAR-10.
Their numbers are stronger than ours on CIFAR-10, but include significantly more inductive bias via a convolutional architecture, so the main takeaway is slightly more relevant towards image tasks rather than arbitrary sequences.

\subsection{Self-attention layers as optimization steps}

The nature of computation performed by self-attention layers has also been explored by other related works.
\cite{bai2019deq} show that a single transformer self-attention block can be trained to perform an optimization step towards finding a stationary point, representing the solution to the task.
\cite{ramsauer2020hopfield} show that the self-attention layer is a gradient step in a Hopfield network with a learning rate of 1, hinting that transformers are capable of storing and retrieving a large amount of patterns with an implicit energy function.
An interesting discussion from \cite{goyal2020inductive} points out a connection in viewing the key-value queries used in attention as similar to function signatures in computer programming: the key maps the input to a type (e.g., float) and the value maps the input to its value (e.g., $3.14$), and if the type matches the function signature, the function can be applied to the value -- this may be particularly relevant when we consider using a single self-attention layer applied to different modalities, as the modality may be embedded in the type.

\subsection{Global workspace theory} \label{sec:gwt}

A common technique for evaluating the embeddings learned by an unsupervised model is to train a linear layer on top of the embeddings for a downstream task \citep{donahue2016bigan, oord2018cpc, chen2020simclr}, which is reasonable when you finetune on the same modality as the pretrained one.
However, when finetuning on a different modality, as in our setting, we have to reframe this notion of generalizable embedding quality -- instead of only finetuning the output layer, we also want to finetune the input layer, and instead evaluate the ability of the frozen intermediate model to perform generalizable \emph{computation}.
This is reminiscent of Global Workspace Theory \citep{baars1993gwt}, which revolves around the notion that there is a ``blackboard'' that different parts of the brain send data to; we might consider the frozen language model as being a blackboard in this setting.
Language might also be a natural choice of model for this blackboard, as there are hypotheses that language may serve as a good multipurpose high-level representation for cognitive behavior and conscious planning \citep{andreas2017l3, goyal2020inductive}.

\subsection{Reservoir computing} \label{sec:resevoir}

Similarly to the FPT setup and Global Workspace Theory, in reservoir computing \citep{tanaka2019reservoir} and echo state networks \citep{jaeger2001echo, jaeger2004harnessing}, a random recurrent network is frozen and only the output readout layer is trained.
These models are very fast to train, using a similar setup as in Section \ref{sec:reservoir}, because the activations of the recurrent network can be cached and it is unnecessary to backpropagate over time.
Somewhat differently to the FPT architecture, echo state networks are recurrent and thus feed back into themselves, which allows the outputs of the random frozen network to modulate future inputs.
Unlike echo state networks, we also notably finetune the input and positional embeddings, which allow the inputs to the frozen network to adapt to a particular modality/for a query to the frozen network to be learned.
Echo state networks are also similar to the perspective of self-attention applying a data-dependent filter to the inputs, as opposed to 1D convolutions, which are fixed filters regardless of the input modality.

\section{Conclusion}

We proposed transferring a pretrained transformer language model for downstream tasks in non-language modalities.
Through extensive empirical evaluation, we showed that these models could achieve performance competitive with transformers fully trained on the downstream task without having to finetune the self-attention and feedforward layers, relying solely on frozen parameters from the language model to perform the bulk of the computation.

We believe this work can serve as the foundation for future work investigating transfer between modalities.
In future, we are interested in investigating the use of other data-rich modalities (e.g., vision) or a hybrid of multiple domains being used to provide the necessary substrate for pretraining a universal computational engine.
It would also be interesting to explore frozen pretrained models for tasks beyond predictive modeling, such as reinforcement learning \citep{abramson2020imitating}.

We note that a limitation of our analysis in that we analyze specific models on a restricted set of tasks.
More investigation can highlight whether or not similar behavior occurs for other models on other tasks.
For instance, in Section \ref{sec:alternative_architectures}, we find the architecture can have a significant impact on results.
As training regimes for these models evolve, performing similar experiments may yield different results, and we are excited for more research in this direction.

For high stakes applications in the real-world, there are potential concerns with transfer of harmful biases from one modality to one another using pretrained transformer models trained on vast quantities of unlabeled, uncurated datasets~\citep{sheng2019woman,bender2021dangers}.
Mitigating these biases is an active area of research~\citep{grover2019bias,choi2020fair}.
Conversely, there are also potential upsides with FPT models being able to better exploit representative datasets from one or more modalities, which merit future investigation as well.

\section*{Acknowledgements}
\addcontentsline{toc}{section}{Acknowledgements}

We would like to thank Luke Metz, Kimin Lee, Fangchen Liu, Roshan Rao, Aravind Srinivas, Nikita Kitaev, Daniel Freeman, Marc'Aurelio Ranzato, Jacob Andreas, and Ashish Vaswani for valuable feedback and discussions.
We would also like to thank members of the community for providing feedback online on an earlier version of this paper.

\clearpage

\section*{Parameter ablations for pretrained models}

\begin{table}[H] 
\begin{center}
\begin{tabular}{c|cccc}
\toprule
\textbf{Task} & \multicolumn{1}{c}{\bf output only} & \multicolumn{1}{c}{\bf output + input} & \multicolumn{1}{c}{\bf output + positions} & \multicolumn{1}{c}{\bf output + layernorm} \\
\midrule
Bit Memory & 76\% & \textbf{98\%} & 93\% & 94\% \\
Bit XOR & 56\% & 72\% & 84\% & \textbf{98\%} \\
ListOps & 15\% & 17\% & 35\% & \textbf{36\%} \\
MNIST & 23\% & 85\% & 93\% & \textbf{96\%} \\
CIFAR-10 & 25\% & 53\% & 38\% & \textbf{54\%} \\
CIFAR-10 LRA & 17\% & 22\% & 30\% & \textbf{39\%} \\
Homology & 2\% &  8\% &  8\% & \textbf{9\%} \\
\bottomrule
\end{tabular}
\end{center}
\caption{Ablation by only finetuning individual types of parameters for pretrained frozen transformers. We bold the most important parameter (measured by highest test accuracy) for each task.}\label{table:finetuning_indep}
\end{table}

\begin{table}[H]
\begin{center}
\begin{tabular}{c|cccc}
\toprule
\textbf{Task} & \multicolumn{1}{c}{\bf output only} & \multicolumn{1}{c}{\bf + layernorm} & \multicolumn{1}{c}{\bf + input} & \multicolumn{1}{c}{\bf + positions} \\
\midrule
Bit Memory & 76\% & 94\% & 100\% & 100\% \\
Bit XOR & 56\% & 98\% & 98\% & 100\% \\
ListOps & 15\% & 36\% & 36\% & 38\% \\
MNIST & 23\% & 96\% & 98\% & 98\% \\
CIFAR-10 & 25\% & 54\% & 60\% & 68\% \\
CIFAR-10 LRA & 17\% & 39\% & 39\% & 39\% \\
Homology & 2\% & 9\% & 10\% & 13\% \\
\bottomrule
\end{tabular}
\end{center}
\caption{Ablation by successively adding certain parameters to the list of finetuned parameters for pretrained frozen transformers.}\label{table:finetuning_add}
\end{table}

\section*{Parameter ablations for random models}

\begin{table}[H] 
\begin{center}
\begin{tabular}{c|cccc}
\toprule
\textbf{Task} & \multicolumn{1}{c}{\bf output only} & \multicolumn{1}{c}{\bf output + input} & \multicolumn{1}{c}{\bf output + positions} & \multicolumn{1}{c}{\bf output + layernorm} \\
\midrule
Bit Memory & 75\% & 75\% & 75\% & 75\% \\
Bit XOR & 50\% & 51\% & 59\% & \textbf{100\%} \\
ListOps & 17\% & 17\% & 18\% & \textbf{35\%} \\
MNIST & 25\% & 28\% & 34\% & \textbf{83\%} \\
CIFAR-10 & 20\% & 24\% & 21\% & \textbf{46\%} \\
CIFAR-10 LRA & 11\% & 16\% & 12\% & \textbf{34\%} \\
Homology & 2\% &  2\% &  6\% & \textbf{9\%} \\
\bottomrule
\end{tabular}
\end{center}
\caption{Finetuning individual types of parameters for random frozen transformers.}\label{table:finetuning_random_indep}
\end{table}

\begin{table}[H]
\begin{center}
\begin{tabular}{c|cccc}
\toprule
\textbf{Task} & \multicolumn{1}{c}{\bf output only} & \multicolumn{1}{c}{\bf + layernorm} & \multicolumn{1}{c}{\bf + input} & \multicolumn{1}{c}{\bf + positions} \\
\midrule
Bit Memory & 75\% & 75\% & 75\% & 76\% \\
Bit XOR & 50\% & 100\% & 100\% & 100\% \\
ListOps & 17\% & 35\% & 36\% & 37\% \\
MNIST & 25\% & 83\% & 92\% & 92\% \\
CIFAR-10 & 20\% & 46\% & 56\% & 62\% \\
CIFAR-10 LRA & 11\% & 34\% & 36\% & 36\% \\
Homology & 2\% & 9\%  & 9\% & 9\% \\
\bottomrule
\end{tabular}
\end{center}
\caption{Ablation by successively adding certain parameters to the list of finetuned parameters for random frozen transformers.}\label{table:finetuning_random_add}
\end{table}

\clearpage

\bibliography{citations}

\begin{thebibliography}{67}
\providecommand{\natexlab}[1]{#1}
\providecommand{\url}[1]{\texttt{#1}}
\expandafter\ifx\csname urlstyle\endcsname\relax
  \providecommand{\doi}[1]{doi: #1}\else
  \providecommand{\doi}{doi: \begingroup \urlstyle{rm}\Url}\fi

\bibitem[Abramson et~al.(2020)Abramson, Ahuja, Brussee, Carnevale, Cassin,
  Clark, Dudzik, Georgiev, Guy, Harley, et~al.]{abramson2020imitating}
Josh Abramson, Arun Ahuja, Arthur Brussee, Federico Carnevale, Mary Cassin,
  Stephen Clark, Andrew Dudzik, Petko Georgiev, Aurelia Guy, Tim Harley, et~al.
\newblock Imitating interactive intelligence.
\newblock \emph{arXiv preprint arXiv:2012.05672}, 2020.

\bibitem[Andreas et~al.(2017)Andreas, Klein, and Levine]{andreas2017l3}
Jacob Andreas, Dan Klein, and Sergey Levine.
\newblock Learning with latent language.
\newblock \emph{arXiv preprint arXiv:1711.00482}, 2017.

\bibitem[Artetxe et~al.(2019)Artetxe, Ruder, and Yogatama]{artetxe2019cross}
Mikel Artetxe, Sebastian Ruder, and Dani Yogatama.
\newblock On the cross-lingual transferability of monolingual representations.
\newblock \emph{arXiv preprint arXiv:1910.11856}, 2019.

\bibitem[Ba et~al.(2016)Ba, Kiros, and Hinton]{ba2016layernorm}
Jimmy~Lei Ba, Jamie~Ryan Kiros, and Geoffrey~E Hinton.
\newblock Layer normalization.
\newblock \emph{arXiv preprint arXiv:1607.06450}, 2016.

\bibitem[Baars(1993)]{baars1993gwt}
Bernard~J Baars.
\newblock \emph{A cognitive theory of consciousness}.
\newblock Cambridge University Press, 1993.

\bibitem[Bai et~al.(2019)Bai, Kolter, and Koltun]{bai2019deq}
Shaojie Bai, J~Zico Kolter, and Vladlen Koltun.
\newblock Deep equilibrium models.
\newblock \emph{arXiv preprint arXiv:1909.01377}, 2019.

\bibitem[Beltagy et~al.(2020)Beltagy, Peters, and Cohan]{beltagy2020longformer}
Iz~Beltagy, Matthew~E Peters, and Arman Cohan.
\newblock Longformer: The long-document transformer.
\newblock \emph{arXiv preprint arXiv:2004.05150}, 2020.

\bibitem[Bender et~al.(2021)Bender, Gebru, McMillan-Major, and
  Shmitchell]{bender2021dangers}
Emily~M Bender, Timnit Gebru, Angelina McMillan-Major, and Shmargaret
  Shmitchell.
\newblock On the dangers of stochastic parrots: Can language models be too big.
\newblock In \emph{Proceedings of the 2020 Conference on Fairness,
  Accountability, and Transparency; Association for Computing Machinery: New
  York, NY, USA}, 2021.

\bibitem[Brown et~al.(2020)Brown, Mann, Ryder, Subbiah, Kaplan, Dhariwal,
  Neelakantan, Shyam, Sastry, Askell, et~al.]{brown2020gpt3}
Tom~B Brown, Benjamin Mann, Nick Ryder, Melanie Subbiah, Jared Kaplan, Prafulla
  Dhariwal, Arvind Neelakantan, Pranav Shyam, Girish Sastry, Amanda Askell,
  et~al.
\newblock Language models are few-shot learners.
\newblock \emph{arXiv preprint arXiv:2005.14165}, 2020.

\bibitem[Chen et~al.(2020{\natexlab{a}})Chen, Radford, Child, Wu, Jun, Luan,
  and Sutskever]{chen2020imagegpt}
Mark Chen, Alec Radford, Rewon Child, Jeffrey Wu, Heewoo Jun, David Luan, and
  Ilya Sutskever.
\newblock Generative pretraining from pixels.
\newblock In \emph{International Conference on Machine Learning}, pp.\
  1691--1703. PMLR, 2020{\natexlab{a}}.

\bibitem[Chen et~al.(2018)Chen, Lucic, Houlsby, and
  Gelly]{chen2018selfmodulation}
Ting Chen, Mario Lucic, Neil Houlsby, and Sylvain Gelly.
\newblock On self modulation for generative adversarial networks.
\newblock \emph{arXiv preprint arXiv:1810.01365}, 2018.

\bibitem[Chen et~al.(2020{\natexlab{b}})Chen, Kornblith, Norouzi, and
  Hinton]{chen2020simclr}
Ting Chen, Simon Kornblith, Mohammad Norouzi, and Geoffrey Hinton.
\newblock A simple framework for contrastive learning of visual
  representations.
\newblock In \emph{International conference on machine learning}, pp.\
  1597--1607. PMLR, 2020{\natexlab{b}}.

\bibitem[Choi et~al.(2020)Choi, Grover, Singh, Shu, and Ermon]{choi2020fair}
Kristy Choi, Aditya Grover, Trisha Singh, Rui Shu, and Stefano Ermon.
\newblock Fair generative modeling via weak supervision.
\newblock In \emph{International Conference on Machine Learning}, pp.\
  1887--1898. PMLR, 2020.

\bibitem[Dai \& Le(2015)Dai and Le]{dai2015semi}
Andrew~M Dai and Quoc~V Le.
\newblock Semi-supervised sequence learning.
\newblock \emph{arXiv preprint arXiv:1511.01432}, 2015.

\bibitem[Deng et~al.(2009)Deng, Dong, Socher, Li, Li, and
  Fei-Fei]{deng2009imagenet}
Jia Deng, Wei Dong, Richard Socher, Li-Jia Li, Kai Li, and Li~Fei-Fei.
\newblock Imagenet: A large-scale hierarchical image database.
\newblock In \emph{2009 IEEE conference on computer vision and pattern
  recognition}, pp.\  248--255. Ieee, 2009.

\bibitem[Devlin et~al.(2018)Devlin, Chang, Lee, and Toutanova]{devlin2019bert}
Jacob Devlin, Ming-Wei Chang, Kenton Lee, and Kristina Toutanova.
\newblock Bert: Pre-training of deep bidirectional transformers for language
  understanding.
\newblock \emph{arXiv preprint arXiv:1810.04805}, 2018.

\bibitem[Donahue et~al.(2016)Donahue, Kr{\"a}henb{\"u}hl, and
  Darrell]{donahue2016bigan}
Jeff Donahue, Philipp Kr{\"a}henb{\"u}hl, and Trevor Darrell.
\newblock Adversarial feature learning.
\newblock \emph{arXiv preprint arXiv:1605.09782}, 2016.

\bibitem[Dosovitskiy et~al.(2020)Dosovitskiy, Beyer, Kolesnikov, Weissenborn,
  Zhai, Unterthiner, Dehghani, Minderer, Heigold, Gelly,
  et~al.]{dosovitskiy2020vit}
Alexey Dosovitskiy, Lucas Beyer, Alexander Kolesnikov, Dirk Weissenborn,
  Xiaohua Zhai, Thomas Unterthiner, Mostafa Dehghani, Matthias Minderer, Georg
  Heigold, Sylvain Gelly, et~al.
\newblock An image is worth 16x16 words: Transformers for image recognition at
  scale.
\newblock \emph{arXiv preprint arXiv:2010.11929}, 2020.

\bibitem[El-Gebali et~al.(2019)El-Gebali, Mistry, Bateman, Eddy, Luciani,
  Potter, Qureshi, Richardson, Salazar, Smart, Sonnhammer, Hirsh, Paladin,
  Piovesan, Tosatto, and Finn]{elgebali2019pfam}
Sara El-Gebali, Jaina Mistry, Alex Bateman, Sean~R Eddy, Aur{\'{e}}lien
  Luciani, Simon~C Potter, Matloob Qureshi, Lorna~J Richardson, Gustavo~A
  Salazar, Alfredo Smart, Erik L~L Sonnhammer, Layla Hirsh, Lisanna Paladin,
  Damiano Piovesan, Silvio C~E Tosatto, and Robert~D Finn.
\newblock {The Pfam protein families database in 2019}.
\newblock \emph{Nucleic Acids Research}, 47\penalty0 (D1):\penalty0 D427--D432,
  2019.
\newblock ISSN 0305-1048.
\newblock \doi{10.1093/nar/gky995}.

\bibitem[Fox et~al.(2013)Fox, Brenner, and Chandonia]{fox2013scop}
Naomi~K Fox, Steven~E Brenner, and John-Marc Chandonia.
\newblock Scope: Structural classification of proteins—extended, integrating
  scop and astral data and classification of new structures.
\newblock \emph{Nucleic acids research}, 42\penalty0 (D1):\penalty0 D304--D309,
  2013.

\bibitem[Frankle et~al.(2020)Frankle, Schwab, and Morcos]{frankle2020batchnorm}
Jonathan Frankle, David~J Schwab, and Ari~S Morcos.
\newblock Training batchnorm and only batchnorm: On the expressive power of
  random features in cnns.
\newblock \emph{arXiv preprint arXiv:2003.00152}, 2020.

\bibitem[Goh et~al.(2021)Goh, Voss, Amodei, Carter, Petrov, Wang, Cammarata,
  and Olah]{goh2021multimodal}
Gabriel Goh, Chelsea Voss, Daniela Amodei, Shan Carter, Michael Petrov,
  Justin~Jay Wang, Nick Cammarata, and Chris Olah.
\newblock Multimodal neurons in artificial neural networks.
\newblock 2021.

\bibitem[Goyal \& Bengio(2020)Goyal and Bengio]{goyal2020inductive}
Anirudh Goyal and Yoshua Bengio.
\newblock Inductive biases for deep learning of higher-level cognition.
\newblock \emph{arXiv preprint arXiv:2011.15091}, 2020.

\bibitem[Grover et~al.(2019)Grover, Song, Agarwal, Tran, Kapoor, Horvitz, and
  Ermon]{grover2019bias}
Aditya Grover, Jiaming Song, Alekh Agarwal, Kenneth Tran, Ashish Kapoor, Eric
  Horvitz, and Stefano Ermon.
\newblock Bias correction of learned generative models using likelihood-free
  importance weighting.
\newblock \emph{arXiv preprint arXiv:1906.09531}, 2019.

\bibitem[He et~al.(2016)He, Zhang, Ren, and Sun]{he2016resnet}
Kaiming He, Xiangyu Zhang, Shaoqing Ren, and Jian Sun.
\newblock Deep residual learning for image recognition.
\newblock In \emph{Proceedings of the IEEE conference on computer vision and
  pattern recognition}, pp.\  770--778, 2016.

\bibitem[Hendrycks \& Gimpel(2016)Hendrycks and Gimpel]{hendrycks2016gelu}
Dan Hendrycks and Kevin Gimpel.
\newblock Gaussian error linear units (gelus).
\newblock \emph{arXiv preprint arXiv:1606.08415}, 2016.

\bibitem[Hernandez et~al.(2021)Hernandez, Kaplan, Henighan, and
  McCandlish]{hernandez2021scaling}
Danny Hernandez, Jared Kaplan, Tom Henighan, and Sam McCandlish.
\newblock Scaling laws for transfer.
\newblock \emph{arXiv preprint arXiv:2102.01293}, 2021.

\bibitem[Hochreiter \& Schmidhuber(1997)Hochreiter and
  Schmidhuber]{hochreiter1997lstm}
Sepp Hochreiter and J{\"u}rgen Schmidhuber.
\newblock Long short-term memory.
\newblock \emph{Neural computation}, 9\penalty0 (8):\penalty0 1735--1780, 1997.

\bibitem[Hou et~al.(2018)Hou, Adhikari, and Cheng]{hou2018deepsf}
Jie Hou, Badri Adhikari, and Jianlin Cheng.
\newblock Deepsf: deep convolutional neural network for mapping protein
  sequences to folds.
\newblock \emph{Bioinformatics}, 34\penalty0 (8):\penalty0 1295--1303, 2018.

\bibitem[Houlsby et~al.(2019)Houlsby, Giurgiu, Jastrzebski, Morrone,
  De~Laroussilhe, Gesmundo, Attariyan, and Gelly]{houlsby2019adapter}
Neil Houlsby, Andrei Giurgiu, Stanislaw Jastrzebski, Bruna Morrone, Quentin
  De~Laroussilhe, Andrea Gesmundo, Mona Attariyan, and Sylvain Gelly.
\newblock Parameter-efficient transfer learning for nlp.
\newblock In \emph{International Conference on Machine Learning}, pp.\
  2790--2799. PMLR, 2019.

\bibitem[Jaeger(2001)]{jaeger2001echo}
Herbert Jaeger.
\newblock The “echo state” approach to analysing and training recurrent
  neural networks-with an erratum note.
\newblock \emph{Bonn, Germany: German National Research Center for Information
  Technology GMD Technical Report}, 148\penalty0 (34):\penalty0 13, 2001.

\bibitem[Jaeger \& Haas(2004)Jaeger and Haas]{jaeger2004harnessing}
Herbert Jaeger and Harald Haas.
\newblock Harnessing nonlinearity: Predicting chaotic systems and saving energy
  in wireless communication.
\newblock \emph{science}, 304\penalty0 (5667):\penalty0 78--80, 2004.

\bibitem[Jumper et~al.(2021)Jumper, Evans, Pritzel, Green, Figurnov,
  Tunyasuvunakool, Ronneberger, Bates, Žídek, Bridgland, Meyer, Kohl,
  Potapenko, Ballard, Cowie, Romera-Paredes, Nikolov, Jain, Adler, Back,
  Petersen, Reiman, Steinegger, Pacholska, Silver, Vinyals, Senior,
  Kavukcuoglu, Kohli, and Hassabis]{jumper2021alphafold}
John Jumper, Richard Evans, Alexander Pritzel, Tim Green, Michael Figurnov,
  Kathryn Tunyasuvunakool, Olaf Ronneberger, Russ Bates, Augustin Žídek, Alex
  Bridgland, Clemens Meyer, Simon A~A Kohl, Anna Potapenko, Andrew~J Ballard,
  Andrew Cowie, Bernardino Romera-Paredes, Stanislav Nikolov, Rishub Jain,
  Jonas Adler, Trevor Back, Stig Petersen, David Reiman, Martin Steinegger,
  Michalina Pacholska, David Silver, Oriol Vinyals, Andrew~W Senior, Koray
  Kavukcuoglu, Pushmeet Kohli, and Demis Hassabis.
\newblock High accuracy protein structure prediction using deep learning.
\newblock 2021.

\bibitem[Kaiser et~al.(2017)Kaiser, Gomez, Shazeer, Vaswani, Parmar, Jones, and
  Uszkoreit]{kaiser2017multitask}
Lukasz Kaiser, Aidan~N Gomez, Noam Shazeer, Ashish Vaswani, Niki Parmar, Llion
  Jones, and Jakob Uszkoreit.
\newblock One model to learn them all.
\newblock \emph{arXiv preprint arXiv:1706.05137}, 2017.

\bibitem[Kiela et~al.(2019)Kiela, Bhooshan, Firooz, Perez, and
  Testuggine]{kiela2019supervised}
Douwe Kiela, Suvrat Bhooshan, Hamed Firooz, Ethan Perez, and Davide Testuggine.
\newblock Supervised multimodal bitransformers for classifying images and text.
\newblock \emph{arXiv preprint arXiv:1909.02950}, 2019.

\bibitem[Kingma \& Ba(2014)Kingma and Ba]{kingma2014adam}
Diederik~P Kingma and Jimmy Ba.
\newblock Adam: A method for stochastic optimization.
\newblock \emph{arXiv preprint arXiv:1412.6980}, 2014.

\bibitem[Krizhevsky et~al.(2009)]{krizhevsky2009cifar}
Alex Krizhevsky et~al.
\newblock Learning multiple layers of features from tiny images.
\newblock 2009.

\bibitem[Li et~al.(2020)Li, Ponti, Vuli{\'c}, and
  Korhonen]{li2020communication}
Yaoyiran Li, Edoardo~M Ponti, Ivan Vuli{\'c}, and Anna Korhonen.
\newblock Emergent communication pretraining for few-shot machine translation.
\newblock \emph{arXiv preprint arXiv:2011.00890}, 2020.

\bibitem[Lu et~al.(2019)Lu, Batra, Parikh, and Lee]{lu2019vilbert}
Jiasen Lu, Dhruv Batra, Devi Parikh, and Stefan Lee.
\newblock Vilbert: Pretraining task-agnostic visiolinguistic representations
  for vision-and-language tasks.
\newblock \emph{arXiv preprint arXiv:1908.02265}, 2019.

\bibitem[Lu et~al.(2020)Lu, Goswami, Rohrbach, Parikh, and
  Lee]{lu2020vilbertmulti}
Jiasen Lu, Vedanuj Goswami, Marcus Rohrbach, Devi Parikh, and Stefan Lee.
\newblock 12-in-1: Multi-task vision and language representation learning.
\newblock In \emph{Proceedings of the IEEE/CVF Conference on Computer Vision
  and Pattern Recognition}, pp.\  10437--10446, 2020.

\bibitem[Miconi et~al.(2018)Miconi, Stanley, and Clune]{miconi2018hebbian}
Thomas Miconi, Kenneth Stanley, and Jeff Clune.
\newblock Differentiable plasticity: training plastic neural networks with
  backpropagation.
\newblock In \emph{International Conference on Machine Learning}, pp.\
  3559--3568. PMLR, 2018.

\bibitem[Oord et~al.(2018)Oord, Li, and Vinyals]{oord2018cpc}
Aaron van~den Oord, Yazhe Li, and Oriol Vinyals.
\newblock Representation learning with contrastive predictive coding.
\newblock \emph{arXiv preprint arXiv:1807.03748}, 2018.

\bibitem[Papadimitriou \& Jurafsky(2020)Papadimitriou and
  Jurafsky]{papadimitriou2020music}
Isabel Papadimitriou and Dan Jurafsky.
\newblock Pretraining on non-linguistic structure as a tool for analyzing
  learning bias in language models.
\newblock \emph{arXiv preprint arXiv:2004.14601}, 2020.

\bibitem[Parisotto et~al.(2020)Parisotto, Song, Rae, Pascanu, Gulcehre,
  Jayakumar, Jaderberg, Kaufman, Clark, Noury,
  et~al.]{parisotto2020stabilizing}
Emilio Parisotto, Francis Song, Jack Rae, Razvan Pascanu, Caglar Gulcehre,
  Siddhant Jayakumar, Max Jaderberg, Raphael~Lopez Kaufman, Aidan Clark, Seb
  Noury, et~al.
\newblock Stabilizing transformers for reinforcement learning.
\newblock In \emph{International Conference on Machine Learning}, pp.\
  7487--7498. PMLR, 2020.

\bibitem[Paszke et~al.(2019)Paszke, Gross, Massa, Lerer, Bradbury, Chanan,
  Killeen, Lin, Gimelshein, Antiga, et~al.]{paszke2019pytorch}
Adam Paszke, Sam Gross, Francisco Massa, Adam Lerer, James Bradbury, Gregory
  Chanan, Trevor Killeen, Zeming Lin, Natalia Gimelshein, Luca Antiga, et~al.
\newblock Pytorch: An imperative style, high-performance deep learning library.
\newblock \emph{arXiv preprint arXiv:1912.01703}, 2019.

\bibitem[Perez et~al.(2018)Perez, Strub, De~Vries, Dumoulin, and
  Courville]{perez2018film}
Ethan Perez, Florian Strub, Harm De~Vries, Vincent Dumoulin, and Aaron
  Courville.
\newblock Film: Visual reasoning with a general conditioning layer.
\newblock In \emph{Proceedings of the AAAI Conference on Artificial
  Intelligence}, volume~32, 2018.

\bibitem[Ponti et~al.(2019)Ponti, Vuli{\'c}, Cotterell, Reichart, and
  Korhonen]{ponti2019towards}
Edoardo~Maria Ponti, Ivan Vuli{\'c}, Ryan Cotterell, Roi Reichart, and Anna
  Korhonen.
\newblock Towards zero-shot language modeling.
\newblock In \emph{Proceedings of the 2019 Conference on Empirical Methods in
  Natural Language Processing and the 9th International Joint Conference on
  Natural Language Processing (EMNLP-IJCNLP)}, pp.\  2893--2903, 2019.

\bibitem[Radford et~al.(2018)Radford, Narasimhan, Salimans, and
  Sutskever]{radford2018gpt}
Alec Radford, Karthik Narasimhan, Tim Salimans, and Ilya Sutskever.
\newblock Improving language understanding by generative pre-training.
\newblock 2018.

\bibitem[Radford et~al.(2019)Radford, Wu, Child, Luan, Amodei, and
  Sutskever]{radford2019gpt2}
Alec Radford, Jeffrey Wu, Rewon Child, David Luan, Dario Amodei, and Ilya
  Sutskever.
\newblock Language models are unsupervised multitask learners.
\newblock 2019.

\bibitem[Radford et~al.(2021)Radford, Kim, Hallacy, Ramesh, Goh, Agarwal,
  Sastry, Askell, Mishkin, Clark, et~al.]{radford2021clip}
Alec Radford, Jong~Wook Kim, Chris Hallacy, Aditya Ramesh, Gabriel Goh,
  Sandhini Agarwal, Girish Sastry, Amanda Askell, Pamela Mishkin, Jack Clark,
  et~al.
\newblock Learning transferable visual models from natural language
  supervision.
\newblock \emph{Image}, 2:\penalty0 T2, 2021.

\bibitem[Raffel et~al.(2019)Raffel, Shazeer, Roberts, Lee, Narang, Matena,
  Zhou, Li, and Liu]{raffel2019t5}
Colin Raffel, Noam Shazeer, Adam Roberts, Katherine Lee, Sharan Narang, Michael
  Matena, Yanqi Zhou, Wei Li, and Peter~J Liu.
\newblock Exploring the limits of transfer learning with a unified text-to-text
  transformer.
\newblock \emph{arXiv preprint arXiv:1910.10683}, 2019.

\bibitem[Ramesh et~al.(2021)Ramesh, Pavolv, Goh, Gray, Chen, Child, Misra,
  Mishkin, Krueger, Agarwal, and Sutskever]{ramesh2021dalle}
Aditya Ramesh, Mikhail Pavolv, Gabriel Goh, Scott Gray, Mark Chen, Rewon Child,
  Vedant Misra, Pamela Mishkin, Gertchen Krueger, Sandhini Agarwal, and Ilya
  Sutskever.
\newblock Dall·e: Creating images from text, 2021.

\bibitem[Ramsauer et~al.(2020)Ramsauer, Sch{\"a}fl, Lehner, Seidl, Widrich,
  Gruber, Holzleitner, Pavlovi{\'c}, Sandve, Greiff,
  et~al.]{ramsauer2020hopfield}
Hubert Ramsauer, Bernhard Sch{\"a}fl, Johannes Lehner, Philipp Seidl, Michael
  Widrich, Lukas Gruber, Markus Holzleitner, Milena Pavlovi{\'c}, Geir~Kjetil
  Sandve, Victor Greiff, et~al.
\newblock Hopfield networks is all you need.
\newblock \emph{arXiv preprint arXiv:2008.02217}, 2020.

\bibitem[Rao et~al.(2019)Rao, Bhattacharya, Thomas, Duan, Chen, Canny, Abbeel,
  and Song]{rap2019tape}
Roshan Rao, Nicholas Bhattacharya, Neil Thomas, Yan Duan, Xi~Chen, John Canny,
  Pieter Abbeel, and Yun~S Song.
\newblock Evaluating protein transfer learning with tape.
\newblock In \emph{Advances in Neural Information Processing Systems}, 2019.

\bibitem[Rao et~al.(2021)Rao, Liu, Verkuil, Meier, Canny, Abbeel, Sercu, and
  Rives]{rao2021msa}
Roshan Rao, Jason Liu, Robert Verkuil, Joshua Meier, John~F. Canny, Pieter
  Abbeel, Tom Sercu, and Alexander Rives.
\newblock Msa transformer.
\newblock \emph{bioRxiv}, 2021.
\newblock \doi{10.1101/2021.02.12.430858}.

\bibitem[Rebuffi et~al.(2017)Rebuffi, Bilen, and Vedaldi]{rebuffi2017adapter}
Sylvestre-Alvise Rebuffi, Hakan Bilen, and Andrea Vedaldi.
\newblock Learning multiple visual domains with residual adapters.
\newblock \emph{arXiv preprint arXiv:1705.08045}, 2017.

\bibitem[Rumelhart et~al.(1985)Rumelhart, Hinton, and
  Williams]{rumelhart1985rnn}
David~E Rumelhart, Geoffrey~E Hinton, and Ronald~J Williams.
\newblock Learning internal representations by error propagation.
\newblock Technical report, California Univ San Diego La Jolla Inst for
  Cognitive Science, 1985.

\bibitem[Sheng et~al.(2019)Sheng, Chang, Natarajan, and Peng]{sheng2019woman}
Emily Sheng, Kai-Wei Chang, Premkumar Natarajan, and Nanyun Peng.
\newblock The woman worked as a babysitter: On biases in language generation.
\newblock \emph{arXiv preprint arXiv:1909.01326}, 2019.

\bibitem[Shin et~al.(2020)Shin, Razeghi, Logan~IV, Wallace, and
  Singh]{shin2020autoprompt}
Taylor Shin, Yasaman Razeghi, Robert~L Logan~IV, Eric Wallace, and Sameer
  Singh.
\newblock Autoprompt: Eliciting knowledge from language models with
  automatically generated prompts.
\newblock \emph{arXiv preprint arXiv:2010.15980}, 2020.

\bibitem[Strodthoff et~al.(2020)Strodthoff, Wagner, Wenzel, and
  Samek]{strodthoff2020udsmprot}
Nils Strodthoff, Patrick Wagner, Markus Wenzel, and Wojciech Samek.
\newblock Udsmprot: universal deep sequence models for protein classification.
\newblock \emph{Bioinformatics}, 36\penalty0 (8):\penalty0 2401--2409, 2020.

\bibitem[Tanaka et~al.(2019)Tanaka, Yamane, H{\'e}roux, Nakane, Kanazawa,
  Takeda, Numata, Nakano, and Hirose]{tanaka2019reservoir}
Gouhei Tanaka, Toshiyuki Yamane, Jean~Benoit H{\'e}roux, Ryosho Nakane, Naoki
  Kanazawa, Seiji Takeda, Hidetoshi Numata, Daiju Nakano, and Akira Hirose.
\newblock Recent advances in physical reservoir computing: A review.
\newblock \emph{Neural Networks}, 115:\penalty0 100--123, 2019.

\bibitem[Tay et~al.(2020)Tay, Dehghani, Abnar, Shen, Bahri, Pham, Rao, Yang,
  Ruder, and Metzler]{tay2020lra}
Yi~Tay, Mostafa Dehghani, Samira Abnar, Yikang Shen, Dara Bahri, Philip Pham,
  Jinfeng Rao, Liu Yang, Sebastian Ruder, and Donald Metzler.
\newblock Long range arena: A benchmark for efficient transformers.
\newblock \emph{arXiv preprint arXiv:2011.04006}, 2020.

\bibitem[Touvron et~al.(2020)Touvron, Cord, Douze, Massa, Sablayrolles, and
  J\'egou]{touvron2020deit}
Hugo Touvron, Matthieu Cord, Matthijs Douze, Francisco Massa, Alexandre
  Sablayrolles, and Herv\'e J\'egou.
\newblock Training data-efficient image transformers \& distillation through
  attention.
\newblock \emph{arXiv preprint arXiv:2012.12877}, 2020.

\bibitem[Vaswani et~al.(2017)Vaswani, Shazeer, Parmar, Uszkoreit, Jones, Gomez,
  Kaiser, and Polosukhin]{vaswani2017attention}
Ashish Vaswani, Noam Shazeer, Niki Parmar, Jakob Uszkoreit, Llion Jones,
  Aidan~N Gomez, Lukasz Kaiser, and Illia Polosukhin.
\newblock Attention is all you need.
\newblock \emph{arXiv preprint arXiv:1706.03762}, 2017.

\bibitem[Wightman(2019)]{wightman2019timm}
Ross Wightman.
\newblock Pytorch image models.
\newblock \url{https://github.com/rwightman/pytorch-image-models}, 2019.

\bibitem[Wolf et~al.(2020)Wolf, Debut, Sanh, Chaumond, Delangue, Moi, Cistac,
  Rault, Louf, Funtowicz, Davison, Shleifer, von Platen, Ma, Jernite, Plu, Xu,
  Scao, Gugger, Drame, Lhoest, and Rush]{wolf2020transformers}
Thomas Wolf, Lysandre Debut, Victor Sanh, Julien Chaumond, Clement Delangue,
  Anthony Moi, Pierric Cistac, Tim Rault, Rémi Louf, Morgan Funtowicz, Joe
  Davison, Sam Shleifer, Patrick von Platen, Clara Ma, Yacine Jernite, Julien
  Plu, Canwen Xu, Teven~Le Scao, Sylvain Gugger, Mariama Drame, Quentin Lhoest,
  and Alexander~M. Rush.
\newblock Transformers: State-of-the-art natural language processing.
\newblock In \emph{Proceedings of the 2020 Conference on Empirical Methods in
  Natural Language Processing: System Demonstrations}, pp.\  38--45, Online,
  October 2020. Association for Computational Linguistics.

\bibitem[Wu et~al.(2021)Wu, Rabe, Li, Ba, Grosse, and Szegedy]{wu2021lime}
Yuhuai Wu, Markus Rabe, Wenda Li, Jimmy Ba, Roger Grosse, and Christian
  Szegedy.
\newblock Lime: Learning inductive bias for primitives of mathematical
  reasoning.
\newblock \emph{arXiv preprint arXiv:2101.06223}, 2021.

\end{thebibliography}
\addcontentsline{toc}{section}{References}

\clearpage

\appendix

\begin{center}{\LARGE{\textbf{Appendix}}}\end{center}
\addcontentsline{toc}{section}{Appendix}

\etocdepthtag.toc{mtappendix}
\etocsettagdepth{mtchapter}{none}
\etocsettagdepth{mtappendix}{subsection}
\tableofcontents

\clearpage

\section{Summary of arXiv Updates}
\label{app:changelog}

We summarize changes made in updated versions:
\begin{enumerate}[label={v\arabic*}.]
    \item (9 Mar 2021) Original release.
    
    \item (30 June 2021) Updated Section \ref{sec:architecture_results} with more analysis of the frozen LSTM architecture and additional experimental details. Added new Section \ref{sec:model_depth} discussing model depth and token mixing, new results in Section \ref{sec:moreparams} discussing how different freezing strategies can improve performance, and attention mask visualization for random frozen transformer to Section \ref{sec:attention_maps}. Included more details about experiments and hyperparameters, and added some new citations (notably \cite{wu2021lime} for related LIME work and \cite{frankle2020batchnorm} for similar frozen analysis for CNNs). Github was also updated to include LSTM architecture, vision pretraining, and remote homology tasks. Minor writing updates.
\end{enumerate}

\section{Background on Transformers}
\label{app:architecture}

In this section, we give a description of the transformer architecture used in our experiments, namely the GPT-2 architecture~\citep{radford2019gpt2}.

\subsection{Self-Attention}

The main subcomponent of the transformer architecture is the self-attention layer, which takes in $l$ input tokens and outputs $l$ output tokens, both of dimension $n_{dim}$.
Each input token $x_i$ is mapped by linear transformations $Q$, $K$, and $V$ -- denoting query, key, and values, respectively -- into $q_i$, $k_i$, and $v_i$.
Both $q_i$ and $k_i$ have dimension $d_k$, and $v_i$ has dimension $d_v$.
To generate the output token $y_i$, dot products are calculated between query $q_i$ and keys $k_j$, and fed into a softmax operation to generate weights $w_i \in [0, 1]$ (in practice, a scaling temperature factor of $\frac{1}{\sqrt{d_k}}$ is used to reduce the sharpness of the softmax).
Then, the weights are used to generate $y_i$ as a weighted sum of all the values, i.e.:
\begin{equation}\label{eq:attention}
    y_i = \sum_{j=1}^l \frac{\text{exp}(q_i^\top k_j)}{\sum_{k=1}^l \text{exp}(q_i^\top k_k)} v_j
\end{equation}

This is extended to \emph{multi-head} attention over $n_{heads}$ heads by doing the above procedure $n_{heads}$ times, and then concatenating.
To recover the original dimension the concatenated vector (of dimension $d_v n_{heads}$) is multiplied by a projection matrix $W_{proj} \in \mathbb{R}^{d_v n_{heads} \times n_{dim}}$.

GPT-2 applies a causal mask to its inputs, i.e. the output token $i$ is only allowed to attend to the input tokens $j \leq i$, which changes the upper bounds of the sums in Equation \ref{eq:attention} to $i$ instead of $l$.
This allows for unsupervised pretraining methods like language modeling (see Appendix \ref{app:objective}).

A residual connection is used to connect the inputs with the outputs of the attention layer.
Then, in the rest of the transformer block, a two-layer MLP is used, conventionally projecting the dimension upwards to $4 \cdot n_{dim}$ for the inner dimension and using the GELU activation function~\citep{hendrycks2016gelu}.
Another residual connection is used to connect the outputs of the MLP with the previous outputs of the attention layer.

This forms the basis of the transformer block.
As it preserves the dimension $n_{dim}$, multiple blocks can be learned and stacked on top of each other $n_{layers}$ times, before feeding the final hidden states to the output layer.
In our work, we only use the output of the last hidden state for classification, although in principle other methods are reasonable.

\subsection{Positional Embeddings}

As the self-attention blocks are permutation-invariant, in order to capture positional information about sequences, positional embeddings are learned.
For each position $i \in (1, \dots, \text{max\_len})$, a vector $p_i$ is learned.
At the front of the transformer, before feeding in the inputs $x_i$ into the self-attention blocks, the positional embeddings are added to the input embeddings as $x_i := x_i + p_i$.

\subsection{Layer Norm}

Layer norm \citep{ba2016layernorm} is frequently used in recurrent and transformer architectures as a means of normalizing the activations.
In particular, for the activations of training example $x$ of dimension $n_{dim}$, it normalizes by the mean and variance over the features:
\begin{equation}
    \tilde{y}_i = \frac{x_i - \text{mean}(\{x_j\}_{j=1}^{n_{dim}})}{\text{std}(\{x_j\}_{j=1}^{n_{dim}})}
\end{equation}

Then, affine scale and shift parameters each of dimension $n_{dim}$ -- $\gamma$ and $\beta$, respectively -- are learned to generate the outputs $y$.
\begin{equation}
    y_i = \gamma_i \tilde{y}_i + \beta_i
\end{equation}

Layer norm is applied twice per self-attention block: once before the attention layer and once before the MLP.
As a result, a total of $4 \cdot n_{layers} \cdot n_{dim}$ layer norm parameters are learned.

\subsection{Pretraining Objective}
\label{app:objective}

GPT-2 is pretrained on an retrogressive language modeling objective optimizing for parameters which maximize the log-likelihood of the data: $\text{max}_\theta \mathbb{E}[\log p_\theta(x)]$.
GPT-2 models sequences autoregressively, factorizing the probability distribution $p(x) = p(x_1, \dots, x_l)$ via chain rule as:
\begin{equation}
    p(x) = \prod_{i=1}^l p(x_i | x_{i-1}, \dots, x_1)
\end{equation}

For the language domain, this objective can be interpreted as ``given the previous $i-1$ words of a sentence, predict the next word''.

\subsection{Model Sizes}

The model sizes from Section \ref{sec:size} are as follows:

\begin{table}[H] 
\begin{center}
\begin{tabular}{c|ccc|c}
\toprule
\textbf{Model Size} & $n_{layers}$ & $n_{dim}$ & $n_{heads}$ & \# Parameters \\
\midrule
Small (Base) & 12 & 768  & 12 & 117M \\
Medium       & 24 & 1024 & 16 & 345M \\
Large        & 36 & 1280 & 20 & 774M \\
\bottomrule
\end{tabular}
\end{center}
\caption{Hyperparameters for architectures for larger model sizes.}\label{table:model_sizes}
\end{table}

The hyperparameters for the experiments with other architectures (Vision Transformer, BERT, Longformer, T5) are the same as for the base model size shown above.

\section{Experimental Details}
\label{app:experimental_details}

We use implementations of and pretrained models from the Huggingface Transformers library \citep{wolf2020transformers}.
We train all models using the Adam \citep{kingma2014adam} optimizer following Pytorch \citep{paszke2019pytorch} defaults.
For all transformer models, we use a learning rate of $10^{-3}$ without learning rate scheduling.
For the remote homology task only, we use a learning rate of $10^{-4}$ as we found it to give better performance than $10^{-3}$.
We generally use the largest batch size that fits on an RTX 2080 Ti graphics card, somewhere between 2 and 16, without gradient accumulation.
Note that except for the remote homology task, we did not tune the FPT hyperparameters.
For all LSTMs, we use a lower learning rate of $3 \times 10^{-4}$ and the same batch sizes as transformers of the same size.
Models are trained to convergence and evaluated on a heldout test set.
\vspace{2em}

\section{Details by Table}

For clarity, we explicitly write out finer details for some experiment sections where numbers can represent different model types.

\subsection{Can pretrained language models transfer to different modalities?}

This section refers to Table \ref{table:main_result} in Section \ref{sec:transfer}.

\textbf{Bit Memory}
\begin{enumerate}
    \item FPT: 12-layer base size FPT model (finetuning input, output, position, and layernorm params).
    \item Full: 12-layer base size GPT-2 model (training all params).
    \item LSTM: 3-layer, 768 hidden dimension LSTM model (training all params).
\end{enumerate}

\textbf{Bit XOR}
\begin{enumerate}
    \item FPT: 12-layer base size FPT model (finetuning input, output, position, and layernorm params).
    \item Full: 12-layer base size GPT-2 model (training all params).
    \item LSTM: 3-layer, 768 hidden dimension LSTM model (training all params).
\end{enumerate}

\textbf{ListOps}
\begin{enumerate}
    \item FPT: 12-layer base size FPT model (finetuning input, output, position, and layernorm params).
    \item Full: number reported from \cite{tay2020lra} (3-layer vanilla transformer).
    \item LSTM: 3-layer, 768 hidden dimension LSTM model (training all params).
\end{enumerate}

\textbf{CIFAR-10}
\begin{enumerate}
    \item FPT: 36-layer large size FPT model (finetuning input, output, position, and layernorm params).
    \item Full: 3-layer, 768 hidden dimension GPT-2 model (training all params).
    \item LSTM: 3-layer, 768 hidden dimension LSTM model (training all params).
\end{enumerate}

\textbf{CIFAR-10 LRA}
\begin{enumerate}
    \item FPT: 12-layer base size FPT model (finetuning input, output, position, and layernorm params).
    \item Full: number reported from \cite{tay2020lra} (3-layer vanilla transformer).
    \item LSTM: 3-layer, 768 hidden dimension LSTM model (training all params).
\end{enumerate}

\textbf{Remote Homology}
\begin{enumerate}
    \item FPT: 12-layer base size FPT model (finetuning input, output, position, and layernorm params).
    \item Full: number reported from \cite{rap2019tape} (12-layer, 512 hidden dimension vanilla transformer).
    \item LSTM: 3-layer, 768 hidden dimension LSTM model (training all params).
\end{enumerate}
\vspace{2em}

\subsection{What is the importance of the pretraining modality?}
\label{app:details_pretraining}

This section refers to Table \ref{table:random} in Section \ref{sec:pretraining}.

\textbf{All tasks}
\begin{enumerate}
    \item FPT: 12-layer base size FPT model (finetuning input, output, position, and layernorm params). This differs from Table \ref{table:main_result}, Section \ref{sec:transfer} only in the CIFAR-10 model size.
    \item Random: 12-layer randomly initialized (default scheme) base size GPT-2 model (training input, output, position, and layernorm params).
    \item Bit: 12-layer base size GPT-2 model (finetuning input, output, position, and layernorm params), after first being fully finetuned on Bit Memory following default random initialization.
    \item ViT: 12-layer, 768 hidden dimension base size ViT model (finetuning input, output, position, and layernorm params), pretrained on 224 $\times$ 224 ImageNet-21k with a patch size of 16. (\texttt{vit\_base\_patch16\_224} from the \texttt{timm} Pytorch library \citep{wightman2019timm}).
    We reinitialize the input layer from scratch to match each task, and do not use a CLS token or an MLP as the output network -- instead using a linear layer from the last token -- matching the protocol for the other methods.
\end{enumerate}

\subsection{How important is the transformer architecture compared to LSTM architecture?}

The following refer to Section \ref{sec:architecture_results}.
In Table \ref{table:random_architecture}:

\textbf{All tasks}
\begin{enumerate}
    \item Trans: 12-layer randomly initialized (default scheme) base size GPT-2 model (training input, output, and layernorm params). Note: same as ``Random'' in Table \ref{table:random}, Section \ref{sec:pretraining}.
    \item LSTM: 3-layer, 768 hidden dimension ``standard'' LSTM (training input, output, and layernorm params). Does not have residual connections or positional embeddings.
    \item LSTM$^*$: 12-layer, 768 hidden dimension LSTM (training input, output, position, and layernorm params). 
\end{enumerate}

Table \ref{table:lstm_layers}:

\textbf{All tasks}
\begin{enumerate}
    \item 12: 12-layer, 768 hidden dimension ``standard'' LSTM (training input, output, and layernorm params).
    \item 3: 3-layer, 768 hidden dimension ``standard'' LSTM (training input, output, and layernorm params).
\end{enumerate}

Table \ref{table:lstm_layers_residual}:

\textbf{All tasks}
\begin{enumerate}
    \item 12-layer LSTM: 12-layer, 768 hidden dimension ``standard'' LSTM (training input, output, and layernorm params). Note: same as ``12'' in Table \ref{table:lstm_layers}, Section \ref{sec:architecture_results}.
    \item + Residual Connections: 12-layer, 768 hidden dimension LSTM with residual connections (training input, output, and layernorm params).
    \item + Positional Embeddings: 12-layer, 768 hidden dimension LSTM with residual connections and positional embeddings (training input, output, position, and layernorm params). Note: same as ``LSTM$^*$'' in Table \ref{table:random_architecture}, Section \ref{sec:architecture_results}.
\end{enumerate}
\vspace{4em}

\subsection{Does language pretraining improve compute efficiency over random initialization?}

This section refers to Table \ref{table:convergence} in Section \ref{sec:compute_efficiency}.

\textbf{All tasks}
\begin{enumerate}
    \item FPT: 12-layer base size FPT model (finetuning input, output, position, and layernorm params). Note: same models as ``FPT'' in Table \ref{table:random}, Section \ref{sec:pretraining}.
    \item Random: 12-layer randomly initialized (default scheme) base size GPT-2 model (training input, output, position, and layernorm params). Note: same models as ``Random'' in Table \ref{table:random}, Section \ref{sec:pretraining}.
\end{enumerate}

\end{document}